\documentclass[lettersize,journal]{IEEEtran}
\usepackage{amsmath,amsfonts}
\usepackage{algorithmic}
\usepackage{algorithm}
\usepackage{array}
\usepackage[caption=false,font=normalsize,labelfont=sf,textfont=sf]{subfig}
\usepackage{textcomp}
\usepackage{stfloats}
\usepackage{url}
\usepackage{verbatim}
\usepackage{graphicx}
\usepackage[numbers]{natbib}
\usepackage{booktabs,multirow,rotating,makecell}
\usepackage{amssymb,amsthm}%
\theoremstyle{plain} 
\newtheorem{theorem}{Theorem}[section]
\newtheorem{lemma}{Lemma}
\newtheorem{proposition}[theorem]{Proposition}
\theoremstyle{definition} 
\newtheorem{definition}{Definition}[section]
\newtheorem{corollary}{Corollary}
\theoremstyle{remark} 

\begin{document}

\title{Estimation of System Parameters Including Repeated Cross-Sectional Data through Emulator-Informed Deep Generative Model}

\author{Hyunwoo Cho*, Sung Woong Cho*, Hyeontae Jo, and Hyung Ju Hwang ~\IEEEmembership{Member,~IEEE,}
\thanks{Hyunwoo Cho is with the Department of Mathematics, Pohang University of Science and Technology, Pohang 37673, Republic of Korea (e-mail:chw51@postech.ac.kr)}
\thanks{Sung Woong Cho is with the Stochastic Analysis and Applied Research Center, Korea Advanced Institute of Science and Technology, Daejeon 34141, Republic of Korea (email:swcho95kr@kaist.ac.kr)}
\thanks{Hyeontae Jo is with the Division of Applied Mathematical Sciences, Korea University, Sejong City 30019, Republic of Korea and Biomedical Mathematics Group, Pioneer Research Center for Mathematical and Computational Sciences, Institute for Basic Science, Daejeon 34126, Republic of Korea (email:korea\_htj@korea.ac.kr)}
\thanks{Hyung Ju Hwang is with AMSquare Corp, Pohang 37673, Republic of Korea, and the Department of Mathematics, Pohang University of Science and Technology, Pohang 37673, Republic of Korea (e-mail:hjhwang@postech.ac.kr)}
\thanks{*These authors contributed equally to this work.} 
\thanks{Manuscript received November 11, 2024. (Co-Corresponding authors: Hyeontae Jo and Hyung Ju Hwang.)}}



\maketitle

\begin{abstract}
Differential equations (DEs) are crucial for modeling the evolution of natural or engineered systems. Traditionally, the parameters in DEs are adjusted to fit data from system observations. However, in fields such as politics, economics, and biology, available data are often independently collected at distinct time points from different subjects (i.e., repeated cross-sectional (RCS) data). Conventional optimization techniques struggle to accurately estimate DE parameters when RCS data exhibit various heterogeneities, leading to a significant loss of information. To address this issue, we propose a new estimation method called the emulator-informed deep-generative model (EIDGM), designed to handle RCS data. Specifically, EIDGM integrates a physics-informed neural network-based emulator that immediately generates DE solutions and a Wasserstein generative adversarial network-based parameter generator that can effectively mimic the RCS data. We evaluated EIDGM on exponential growth, logistic population models, and the Lorenz system, demonstrating its superior ability to accurately capture parameter distributions. Additionally, we applied EIDGM to an experimental dataset of Amyloid beta 40 and beta 42, successfully capturing diverse parameter distribution shapes. This shows that EIDGM can be applied to model a wide range of systems and extended to uncover the operating principles of systems based on limited data.
\end{abstract}


\section{Introduction}\label{sec:intro}
A system of differential equations (DE) is essential for modeling the dynamics of various systems, offering scientific and mechanistic insights into physical and biological phenomena. The solutions to a DE largely depend on its parameters and determining these parameters is crucial for fitting the solutions to observed data. Specifically, the distribution of the estimates provides additional insights, such as uncertainty quantification of data or heterogeneity of underlying dynamics \citep{jo2024density}, leading to a more comprehensive understanding of the phenomena. These estimation tasks are typically performed using optimization methods that update parameters to make the corresponding solutions of the system more closely match the observational data, especially when the data are obtained by observing individual samples (subjects) over time (i.e., time series data). 
Different from time series data, in various fields, such as biology, economics, and political science, data are often collected from different samples or groups of individuals at multiple points in time (i.e., repeated cross-sectional (RCS) data \citep{beck2007random,beck2011modeling,Pan2019, stevens2018hbsc, bryman2016social}). For instance, \cite{jeong2022systematic} analyzed sequential data on PER protein levels in fruit flies to study the dependence of molecular characteristics on neurons. However, obtaining PER levels at different time points entailed the death of the flies, limiting continuous data collection over time. In another study, \cite{lundsten2020radiosensitizer} used an exponential growth model to investigate how drugs affected tumor sizes in mice over time. As the study progressed, mice were sacrificed, complicating the association of observational data and resulting in RCS data. Furthermore, RCS data can be obtained from community surveys reflecting the change of opinions from different people over time (e.g., opinion polls by Gallup, the Michigan Consumer Sentiment Index, records of Congressional votes, Supreme Court decisions, and presidential statements \citep{clarke2005men, hopkins2012whose, lebo2007strategic, segal2002supreme, wood2009myth}).

While fitting parameters with time-series data is feasible with classical optimization methods, handling RCS data poses a challenge. Specifically, \cite{jo2024estimating} demonstrated that estimating parameters using the mean value of RCS data (e.g., \cite{jeong2022systematic}) or a Gaussian process (GP)-based model calibration (e.g., \cite{smith2018host, chung2019parameter, zhang2020improved, binois2021hetgp}) yields significant mismatches, resulting in incorrect interpretation of a given phenomenon. As GP-based model calibration only depends on the mean and covariance of data at each time point, it fails to capture the complete information contained within RCS data, yielding only unimodal distribution estimates. Bayesian methods, including approximate Bayesian computation (ABC) \cite{marjoram2003markov, cranmer2020frontier} and Metropolis-Hastings (MH) algorithms \cite{metropolis1953equation, hastings1970monte}, also struggle to accurately estimate parameter distributions, due to their sensitivity to the prior distribution \cite{swigon2019importance}. To improve their applicability, \cite{jo2024estimating} developed a method to estimate parameters while preserving the information contained in RCS data. However, this method is efficient only when the number of observations in the RCS data is small because this method requires large computational costs for many artificial trajectories.

In this paper, we introduce a novel approach for inferring the system parameters from large RCS data, called the emulator-informed deep generative model (EIDGM), which integrates both a hyper physics-informed neural networks (HyperPINN, \cite{de2021hyperpinn}), and Wasserstein generative adversarial network (WGAN, \cite{arjovsky2017wasserstein}) (Fig. \ref{fig:1}) in the following manner: 1) HyperPINN generates solutions across varying parameters, while 2) WGAN selects solutions that best fit the given RCS data. We then conducted validation tasks to estimate the true parameter distributions with the following mathematical models: 1) exponential growth, 2) logistic population models \citep{yada2023few}, and 3) the Lorenz system \citep{chung2019parameter}. Through this validation, we show that EIDGM can accurately estimate the shape of the parameter distribution. Next, we also applied EIDGM to a real-world dataset (biomarkers for diagnosing dementia), showing that implicit patterns of biomarkers over time can be grouped per individual. These results suggest that EIDGM enables a more comprehensive, deeper, and improved understanding of the RCS data. By analyzing the shape of these parameter distributions, we expect that EIDGM can provide an opportunity to better understand dynamical systems.

\subsection{Contribution}
\begin{itemize}
\item We highlight the significance of incorporating deep learning methods into the traditional analysis of dynamics in natural or engineered systems. By leveraging the efficiency of deep learning in both emulating numerical solvers (HyperPINN) and estimating distributions (WGAN), this study demonstrates how complex parameter estimation tasks can be performed more efficiently and accurately.

\item  EIDGM overcomes the limitations of traditional estimation methods for RCS data, such as a GP-based model calibration. These conventional approaches often result in significant mismatches and inaccurate parameter interpretations. In contrast, EIDGM provides more precise and reliable parameter estimation.

\item Through the precise estimation of parameters, EIDGM provides a more comprehensive, deeper understanding of dynamical systems. This contribution is significant for research in various fields such as economics, political science, and biology, where RCS data are prevalent. In other words, EIDGM offers new insights and opportunities for future research in understanding complex dynamical systems.
\end{itemize}

\subsection{Related works}

\textbf{Development of DE solver using neural networks} Deep learning algorithms have been applied to solving DEs. \cite{yu2018deep} utilized neural networks with a variational formulation in the loss function to solve high-dimensional DEs. However, this approach yields a complicated loss landscape, causing the neural networks to converge to a local minimum. To overcome this, \cite{raissi2019physics} introduced Physics-informed neural networks (PINN) by minimizing a residual loss function that directly measures how well the neural networks satisfy DE. Through PINN, the neural networks can fit the given data while satisfying DEs as constraints.

While PINN can be utilized when the set of parameters is specifically given, it requires a high computational cost for training with different sets of parameters. Thus, PINN is not efficient for constructing an emulator that immediately generates a DE solution. To reduce such computational costs, PINN incorporating the embedding method has been developed by adding parameters as input to the neural network \citep{park2019deepsdf, chen2019learning, mescheder2019occupancy}. Similarly, a lot of variations of the emulator for a DE solver have been proposed (e.g., DeepONet \citep{lu2021learning, hadorn2022shift, seidman2022nomad}). However, these methods still require large neural networks and more training time. 

More recently, PINN incorporating hypernetworks \cite{ha2016hypernetworks} has been developed \citep{de2021hyperpinn}. Specifically, given a set of parameters, the hypernetworks generate the weights for the main neural network. The main network can immediately calculate DE solutions for various sets of parameters. Later, theoretical evidence showed that hypernetworks have greater expressivity in spanning the solution space compared to embedding methods \citep{galanti2020modularity, lee2023hyperdeeponet}. Therefore, we use the hypernetwork structure for the emulator, efficiently generating DE solutions with various sets of parameters.

\textbf{Estimation of probability distribution using GAN} GANs have been proposed to create fake data that are indistinguishable from real data \citep{goodfellow2014generative}. Due to their applicability, GANs are widely used to estimate complex probability 
distributions in various areas, including uncertainty quantification for parameters or initial conditions of DEs. Specifically, \cite{kadeethum2021framework, patel2022solution} utilized a GAN to find the relationship between parameters and solutions of a given DE. Later, \cite{kadeethum2021framework} modified the conditional GAN (cGAN) architecture \citep{mirza2014conditional} to infer parameters from each real data sample. While this approach is feasible for given time-series data, it is not applicable to handling RCS data, as two consecutive observation points in the RCS cannot be connected. Therefore, we modified the Wasserstein GAN (WGAN) with a HyperPINN-based emulator.

\textbf{Statistical inference of parameters over simulations}
The development of parameter estimation methods that integrate (numerical) simulators and generative models has advanced significantly, particularly within the domain of simulation-based inference (SBI). SBI involves performing statistical inference on probabilistic models using simulation samples $x$ derived from underlying parameters $\mathbf{p}$, where the likelihood $p(x|\mathbf{p})$ is not explicitly given. For example, Ramesh \textit{et al.} combined both GAN and a numerical solver \citep{ramesh2022gatsbi}. These two parts were used for simulating samples and approximating the posterior distribution of underlying parameters, respectively. More recently, approaches that incorporate transformers (\cite{vaswani2017attention}) and probabilistic diffusion models (\cite{song2020denoising}) have been proposed to handle missing or unstructured data \citep{gloeckler2024all}. However, these models lack a specific design for RCS data, leaving uncertainty about their ability to accurately approximate posterior distributions $p(\mathbf{p}|x)$ in the presence of significant heterogeneity.

\section{Methods}\label{sec:method}

\subsection{Description of problems in estimating parameters of DEs with RCS data}\label{subsec:rcs data}

We propose a method for estimating the distribution of parameters within a time-evolutionary ordinary differential equation (ODE), represented as
\begin{equation} \label{equation:ODE}
\mathbf{y}'(t) = f[\mathbf{y}(t), \mathbf{p}],
\end{equation}
where $\mathbf{y}(t;\mathbf{p}) \in \mathbb{R}^{n_{y}}$ denotes the solution with dimension $n_y$ at time $t$. The set of parameters $\mathbf{p}\in\mathbb{R}^{n_p}$ represents biological or physical properties (such as growth rate or carrying capacity). Here, we aim to estimate the posterior distribution of the parameter $\mathbf{p}$ that can fit the corresponding solution $\mathbf{y}(t;\mathbf{p})$ to the RCS data $Y$. Specifically, $Y$ consists of the data points $(t_{r},\mathbf{y}_{j_{r}})$, where $\{t_{r}\}_{r=1}^{T}$ denotes the $T$ observation time points and $j_{r} \in \{1,2,\dots,J_{r}\}$ is an index with varying maximum $J_{r}$ depending on $r$ for the given RCS data.

\subsection{HyperPINN as an emulator for DE solver}
In this section, we build an emulator that can immediately provide the solution $\mathbf{y}(t)$ of Eq.~(\ref{equation:ODE}) given $\mathbf{p}$. The structure of the emulator is motivated by HyperPINN (\cite{de2021hyperpinn}), which uses two fully connected neural networks: a hypernetwork $h$ (\cite{ha2016hypernetworks, galanti2020modularity, lee2023hyperdeeponet}) and fully connected neural network $m$ (main network). Specifically, the structure of $h$, with weights and biases $\theta_h$, is designed to map the set of parameters $\mathbf{p}$ to the weights and biases of the main network $m$, $\theta_{m}$ (detailed nodes and activation functions are provided in Table. \ref{table:network settings}): 
\begin{equation} \label{equation:hypernetwork}
\theta_{m}(\mathbf{p})=h(\mathbf{p};\theta_h).
\end{equation} 
Once $\theta_{m}(\mathbf{p})$ is obtained by training the hypernetwork, these values are used as the weights and biases of the main network $m$. That is, the output of the main network $m$ is directly determined by the outputs of the hypernetwork. As a result, the main network immediately produces a function $m(t;\theta_m(\mathbf{p}))$ that closely approximates the solution of Eq.~(\ref{equation:ODE}), $\mathbf{y}(t;\mathbf{p})$:
\begin{equation}\label{equation:mainnetwork}
\mathbf{y}(t;\mathbf{p}) \approx m(t;\theta_m(\mathbf{p})).
\end{equation}

To train the HyperPINN for the task described above, we first define the probability distribution of parameters $\mathbf{p}$, denoted by $\mathcal{D}$, as well as the time interval $[t_1, t_T]$, which encompasses the period covered by the experimental data. Next, we construct two loss functions based on \citep{de2021hyperpinn}: 1) data loss $L_{\text{data}}$ and 2) physics loss $L_{\text{physics}}$. First, $L_{\text{data}}$ is used to fit the output of the main network $m$ to the solution $\mathbf{y}(t;\mathbf{p})$ by minimizing their differences:
\begin{equation}\label{equation:loss_data}
L_{\text{data}}(\theta_h) = \sum_{i=1}^{T_{obs}} \mathbb{E}_{\mathbf{p}\sim\mathcal{D}} \left|m(t^o_{i}; \theta_m(\mathbf{p})) - \mathbf{y}(t^o_{i};\mathbf{p})\right|^2,
\end{equation}
where $\{t^{o}_{i}\}_{i=1}^{T_{obs}}$ denotes the $T_{obs}$ observation time points from the experimental data. Next, we introduce physics loss, which measures how well the output of the main network $m$ satisfies the DE of Eq.~(\ref{equation:ODE}), where $\mathbb{E}$ represents the expectation over the probability distribution $\mathcal{D}$. The measurement can be quantified by substituting the output of the main network into the DE:

\begin{multline}\label{equation:loss_phy}
L_{\text{physics}}(\theta_h) = \sum_{i=1}^{T_{col}} \mathbb{E}_{\mathbf{p}\sim\mathcal{D}} 
\Bigg| \frac{d}{dt} m(t_i^{c}; \theta_m(\mathbf{p})) \\
- f\big[m(t_i^{c}; \theta_m(\mathbf{p})), \mathbf{p}_{j}, t_{i}^{c} \big] \Bigg|^2,
\end{multline}

where $\{t_i^{c}\}_{i=1}^{T_{col}}$ represents the collocation time points within the time interval $[t_1, t_T]$. By minimizing Eqs.~(\ref{equation:loss_data},\ref{equation:loss_phy}), we expect that the output of the main network $m(t;\theta_m(\mathbf{p}))$ will not only closely approximate $\mathbf{y}(t;\mathbf{p})$ but also accurately describe the underlying dynamics of the system. We also provide a detailed mathematical analysis for the training framework in the Supplemental materials. While the loss functions of Eq.~(\ref{equation:loss_data}-\ref{equation:loss_phy}) are useful for understanding the training frameworks theoretically, they are not efficient for implementation on computational devices. Hence, in this study, we employed discretized versions of the two loss functions of Eq.~(\ref{equation:loss_data}-\ref{equation:loss_phy}) as follows: 

\begin{align}
    L_{\text{data(disc)}}(\theta_h) & = \sum_{i=1}^{T_{obs}} \sum_{j=1}^{N_p}
    \left|m(t^{o}_{i}; \theta_m(\mathbf{p}_j)) - \mathbf{y}(t^{o}_{i};\mathbf{p}_j)\right|^2, \label{equation:loss_data(disc)}
\end{align}
\begin{multline}\label{equation:loss_phy(disc)}
    L_{\text{physics(disc)}}(\theta_h) = \sum_{i=1}^{T_{col}} \sum_{j=1}^{N_p}
    \Bigg| \frac{d}{dt} m(t^{c}_{i}; \theta_m(\mathbf{p}_j)) \\
    - f[m(t^{c}_{i}; \theta_m(\mathbf{p}_j)), \mathbf{p}_{j}, t^{c}_{i}] \Bigg|^2.
\end{multline}
where $\{\mathbf{p}_j\}_{j=1}^{N_p}$ denotes the set of $N_p$ parameters sampled from the probability distribution $\mathcal{D}$. $\mathbf{y}(t;\mathbf{p})$ of Eq.~(\ref{equation:ODE}) with different $\mathbf{p}$ can be obtained through a DE solver (e.g., LSODA in \textit{Scipy} package). The choice of $N_p$ depends on the number of time points $T$, number of nodes, and type of activation function in the HyperPINN. We refer the reader to Theorems \ref{appendix:theorem_interpolation(one timepoint)} and \ref{appendix:theorem_interpolation} in the supplemental materials, where detailed analyses, including error analysis, are provided.
Finally, we assign weights $\alpha, \beta$ to the two discretized loss functions, Eq.~(\ref{equation:loss_data(disc)}-\ref{equation:loss_phy(disc)}), to prevent biases arising from the initial values of the loss functions. The units of the two loss functions are not equal in general. Therefore, the training can be conducted by minimizing the total loss function, $L(\theta_h)$, which is defined as the sum of the two loss functions with positive weights $\alpha$ and $\beta$:
\begin{equation}\nonumber
L(\theta_h) = \alpha L_{\text{data}}(\theta_h) + \beta L_{\text{physics(disc)}}(\theta_h).
\end{equation}

\subsection{WGAN framework for estimating parameter distribution}
In this section, we aim to find the parameter distribution $\mathbf{p}$ that generates the given RCS data $Y$. For this task, we obatin an initial guess of $\pi (\mathbf{p})$ and sample $N$ parameters $\{\mathbf{p}_{i}\}_{i=1}^{N}$. For each $\mathbf{p}_{i}$, we can simultaneously obtain approximations for solutions of Eq.~(\ref{equation:ODE}) through HyperPINN as follows:
\begin{equation}\nonumber
\tilde{Y} = \{\{(t_{r},\mathbf{\tilde{y}}_{i,r})\}_{r=1}^{T}\}_{i=1}^{N} = \{\{(t_{r},m(t_{r};\mathbf{p}_{i})) \}_{r=1}^{T}\}_{i=1}^{N}.
\end{equation} Next, we adjust the parameter distribution $\pi(\mathbf{p})$ so that the distribution of $\tilde{Y}$ is sufficiently close to that of the given RCS dataset $Y = \{(t_{r},\mathbf{y}_{j_{r}} )\}_{r=1}^{T}$. This adjustment involves measuring the difference between $\tilde{Y}$ and $Y$ and then modifying the parameter set $\{ \mathbf{p}_{i}\}_{i=1}^{N}$ to minimize this difference.

For this task, we employ WGAN with gradient penalty, providing a correct parameter distribution through a generator \citep{arjovsky2017wasserstein, gulrajani2017improved}. Using WGAN, we aim to reduce the following Wasserstein distance (with Kantorovich–Rubinstein duality \citep{villani2009optimal}) between the distributions of $\tilde{Y}$ and $Y$ , denoted as $\mu_{\tilde{Y}}$ and $\mu_{Y}$, respectively: 
\begin{multline}\label{w_dist}
d(\mu_{\tilde{Y}}, \mu_{Y}) = 
\sup_{\{f \in Lip(\mathbb{R}^{d},\mathbb{R}) : \| f\|_{Lip}\leq1\}} 
\mathbb{E}_{(\tilde{t},\mathbf{\tilde{y}} ) \sim \mu_{\tilde{Y}}}[f(\tilde{t},\mathbf{\tilde{y}} )] \\
- \mathbb{E}_{(t,\mathbf{y} ) \sim \mu_{Y}}[f(t,\mathbf{y} )],
\end{multline}
where $\mathbb{E}$ denotes the expectation, and $Lip(\mathbb{R}^{d},\mathbb{R})$ represents the set of all real-valued 1-Lipschitz functions on $\mathbb{R}^d$ (i.e., $Lip(\mathbb{R}^{d},\mathbb{R})=\{f:\mathbb{R}^{d}\rightarrow\mathbb{R} | \|f\|_{Lip}=\sup_{x \neq y } \frac{|f(x)-f(y)|}{|x-y|}<\infty $\}).

For each iteration stage of WGAN, we first sample a set of latent variables $\{ \mathbf{z}_{i} \}_{i=1}^{N}$ from the standard normal distribution, $\mathcal{N}(0,1)$. The generator $G(\cdot ;\theta_{G})$ in WGAN, with weights and biases $\theta_{G}$, maps $\{ \mathbf{z}_{i} \}_{i=1}^{N}$ to a set of parameters $\{ \mathbf{p}_{i}\}_{i=1}^{N} = \{ G(\mathbf{z}_{i} ;\theta_{G})\}_{i=1}^{N}$. Then, we immediately obtain $\tilde{Y}$ corresponding to $\{ \mathbf{p}_{i} \}_{i=1}^{N}$ through the emulator in Eq.~(\ref{equation:hypernetwork}-\ref{equation:mainnetwork}). Then, the discriminator $D(\cdot;\theta_{D})$ in WGAN, with weights and biases $\theta_{D}$, calculates the loss $L_{D}(\theta_{G},\theta_{D})$ (i.e., the difference between $\tilde{Y}$ and $Y$) to minimize 
\begin{align}
\displaystyle L_{D}(\theta_{G},\theta_{D}) & = 
-\frac{1}{N}\sum_{k=1}^{|\tilde{Y}|}D(t_{k}, \mathbf{y}_{k};\theta_{D}) \nonumber \\
& \quad + \frac{1}{N}\sum_{k=1}^{|Y|}D(\tilde{t}_{k},\mathbf{\tilde{y}}_{k};\theta_{D}) \nonumber \\
& \quad + \frac{\lambda }{N}\sum_{k=1}^{|Y|}(||\nabla_{(\hat{t}_{k},\mathbf{\hat{y}}_{k})}D(\hat{t}_{k},\mathbf{\hat{y}}_{k};\theta_{D})||_{2}-1)^{2}, \nonumber
\end{align}
where the augmented data $\displaystyle (\hat{t}_{k},\mathbf{\hat{y}}_{k})$ for calculating the gradient penalty (in the last term of $\displaystyle L_{D}(\theta_{G},\theta_{D})$) are defined as
$$
\displaystyle (\hat{t}_{k},\mathbf{\hat{y}}_{k}) = \epsilon_{k} (\tilde{t}_{k}, \mathbf{\tilde{y}}_{k}) + (1-\epsilon_{k}) (t_{k}, \mathbf{y}_{k})
$$
with uniform random coefficient $\epsilon_{k} \sim U[0,1]$ for each $k \in \{1,2,\dots,|Y|\}$. Note that when $|\tilde{Y}| > |Y|$, the subset of $\tilde{Y}$ with size $|Y|$ for calculating the gradient penalty is chosen randomly for each iteration. The gradient penalty term enforces the discriminator $D$ to be a 1-Lipschitz function, ensuring that minimizing the loss is equivalent to finding $f$ in Eq.~(\ref{w_dist})(\citep{gulrajani2017improved}). Following the recommendations in \citep{gulrajani2017improved}, we set the coefficient $\lambda$ to 10.

For each generator and discriminator in WGAN, we employed fully connected neural networks with hyperbolic tangent functions as the non-linear activation functions (also see the hyperparameters for WGAN in Table \ref{table:network settings}). Given the scarcity of data, especially when the underlying parameter distribution is multi-modal, mini-batch training can introduce significant instability. Therefore, we utilized full-batch training, generating the same volume of fake data as real data to enhance stability.
\begin{figure*}[!t]
    \centering
    \includegraphics[width=\textwidth]{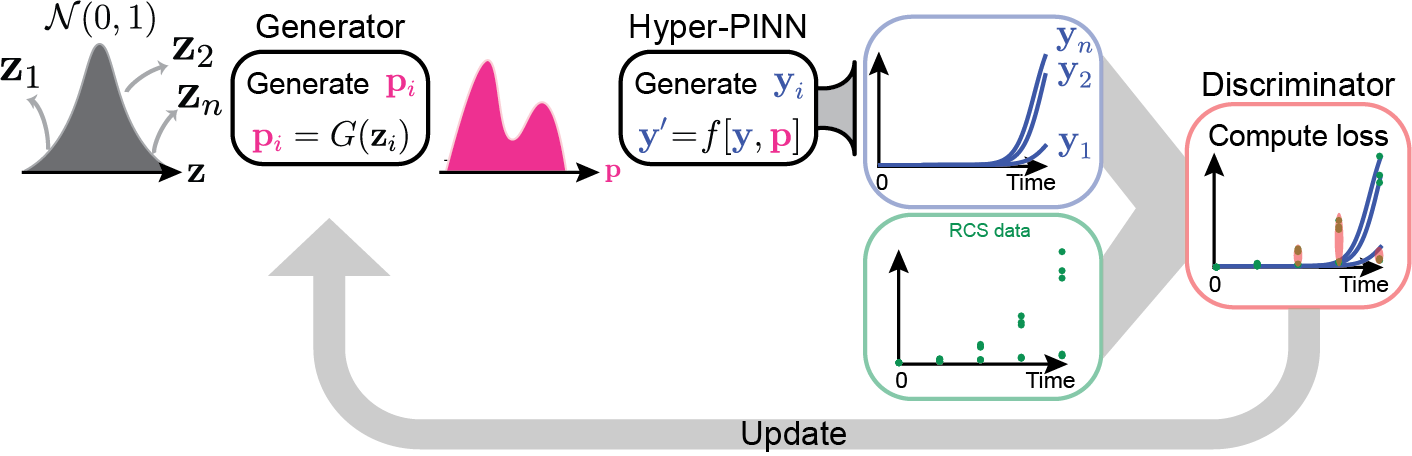}
    \caption{{\bf Schematic of Emulator-Informed Deep-Generative Model (EIDGM)} A generator in WGAN is trained to produce a possible set of parameters. These parameters are used as inputs of HyperPINN, immediately producing solutions corresponding to the set of parameters. We then calculate the loss through the discriminator in WGAN by measuring the difference (distribution) between solutions and RCS data. By minimizing the loss, the generator produces the precise parameter distributions of the DE.}
    \label{fig:1}
\end{figure*}
\section{Results}
\subsection{Development of Emulator-Informed Deep Generative Mo\-del (EIDGM)}

The EIDGM operates in the following stages: 1) We build an emulator using HyperPINN that immediately produces the solution of the DEs corresponding to the given set of parameters. 2) Next, we randomly sample $N$ sets of parameters through a generator of WGAN and produce corresponding solutions through the emulator (Fig. \ref{fig:1}, Generator, Hyper-PINN). 3) Then, we measure the difference between the solutions and given RCS data via a discriminator in WGAN (Fig. \ref{fig:1}, Discriminator). 4) By minimizing the difference and updating both the discriminator and generator, we show that EIDGM accurately captures true parameter distributions. 
To evaluate the efficacy of EIDGM in estimating parameter distributions, we used four different time-evolutionary DEs: an exponential growth model, a logistic population model \citep{yada2023few}, and the Lorenz system. These problems demonstrate that EIDGM can accurately estimate true parameter distributions and predict system behaviors even in the presence of data heterogeneity. Next, we compared estimation performance using different types of emulators with GP \citep{chung2019parameter}, DeepONet \citep{lu2019deeponet}, and EIDGM. Among the emulators, HyperPINN generally provides accurate and precise parameter estimates (Table. \ref{table:comparison}). Detailed test procedures are provided below.

\begin{table}\centering
\caption{{\bf Accuracy of parameter estimates from three different models.} We quantified the accuracy of the estimates using the sum of one-dimensional Wasserstein distances between the parameter distributions for each projection. \textbf{Bold} font indicates the lowest value among the three models.}

\begin{tabular*}{0.48\textwidth}{@{\extracolsep{\fill}} *{5}{c} }
\toprule 
\multicolumn{2}{c}{\multirow{2.2}{*}{Experiments}} &
\multicolumn{3}{c}{Models}  \\
\cmidrule{3-5}
& & GP & DeepONet+WGAN & EIDGM \\ 
\midrule
\multirow{3.2}{*}{Exponential} & uni-modal & 5.60e-01 & 1.72e-02 & \textbf{4.00e-03} \\ & bi-modal & 9.32e-01 & 4.07e-02  & \textbf{3.69e-02}  \\ & tri-modal & 7.35e-01  & \textbf{4.01e-02}  & 5.31e-02  \\
\addlinespace\midrule
\multirow{3.2}{*}{Logistic} & uni-modal & 4.9e-01 & 5.63e-01 & \textbf{3.01e-02} \\ & bi-modal & 1.56e00 & 1.01e00 & \textbf{2.09e-01} \\ & tri-modal & 1.44e00 & 5.69e-01 & \textbf{1.14e-01} \\
\addlinespace\midrule
\multirow{3.2}{*}{Lorenz} & uni-modal & 6.51e-01 & 3.76e-01 & \textbf{1.36e-01} \\ & bi-modal & 4.53e00 & 6.41e-01 & \textbf{3.91e-01} \\ & tri-modal & 3.82e00 & 4.43e-01 & \textbf{3.43e-01} \\
\bottomrule
\end{tabular*}\label{table:comparison}
\end{table}

\subsection{Exponential growth model}
\begin{figure}[ht!]
    \centering
    \includegraphics[width=0.48\textwidth]{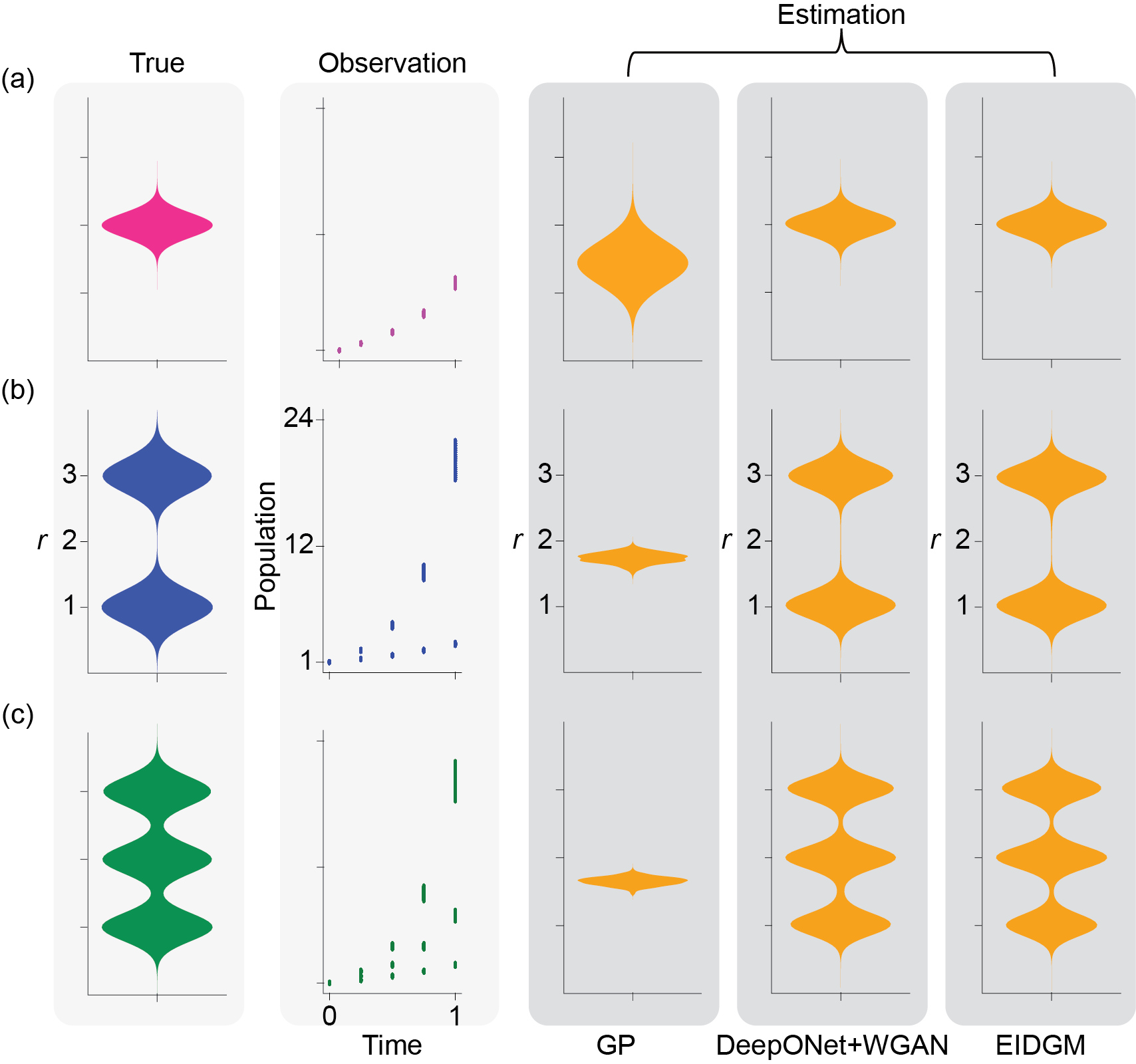}
   \caption{{\bf Visualization of parameter estimates for exponential growth model with three different emulators: Gaussian Process (GP), DeepONet (with WGAN), and EIDGM (HyperPINN with WGAN).}(a) If data are obtained from an exponential model with one parameter peak (RCS Data), all methods accurately estimate the underlying parameters (Right, Estimation). (b), (c) When data are obtained with multiple parameter peaks, GP fails to estimate the underlying parameter distribution, while the Emulator+WGAN models accurately estimate the parameters.}
   \label{fig:exp}
\end{figure}

The exponential growth model describes changes in population size $y(t)$ over time $t$:
\begin{equation*}
y' = ry,
\end{equation*}
where $r$ represents the population growth rate. We first obtain a distribution of parameters with a single peak at $r=2$ (Fig. \ref{fig:exp}(a), True). After sampling $36$ parameter values from the underlying distribution, we generate snapshots for each time $t= 0, 0.25, 0.5, 0.75, 1$, which are from the trajectories generated with $36$ sampled parameters as the RCS dataset (Fig. \ref{fig:exp}(a), Observation) (also see \textit{Simulation dataset generation} in the Supplemental materials for details). We next estimate parameter distributions corresponding to the RCS dataset using three different emulators: GP, DeepONet (with WGAN), and EIDGM (Fig. \ref{fig:exp}(a), Estimation, red). In this case, all three methods can accurately estimate the underlying distribution. Unlike with a single peak, only DeepONet and EIDGM can accurately estimate the parameter distributions when the underlying distributions have different peaks (Fig. \ref{fig:exp}(b-c)). We also provide quantified results using the Wasserstein distance, which measures the distance between true and estimated distributions, in Table. \ref{table:comparison}. 

\subsection{Logistic population model}
\begin{figure}[h]
    \centering
    \includegraphics[width=0.48\textwidth]{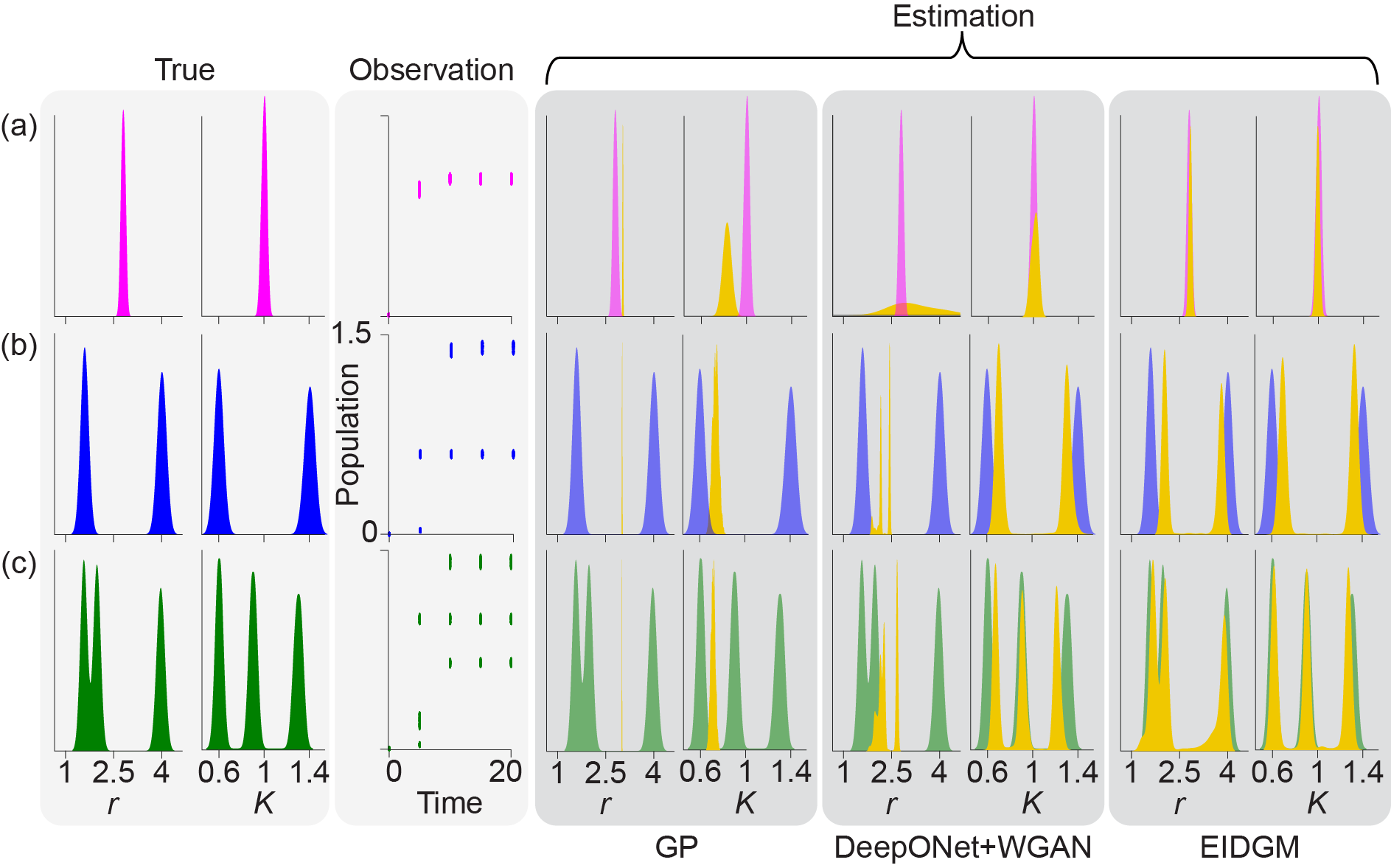}
    \caption{{\bf Visualization of estimation results for logistic population model with three different emulators: GP, DeepONet, and HyperPINN+WGAN (EIDGM).}(a) If data are obtained from a logistic model with one parameter peak (Left, Observation), all methods accurately estimate the underlying parameters (Right, Estimation). (b) (c) Only EIDGM accurately estimates the parameters when there are two or three parameter peaks.}
    \label{log1}
\end{figure}

The logistic population model represents the changes in population size $y(t)$ over time $t$ with the maximum population size $K$:
\begin{equation*}
y' = ry(1-y/K).
\end{equation*}
We begin by constructing distributions for $r$ and $K$ with a single peak, respectively (Fig. \ref{log1}(a), True). Note that the peak values are derived from ranges that were estimated in previous studies (\cite{yada2023few, whittington2018spatiotemporal, hao2022optimal}) (also see Table \ref{table:comparison} for detailed parameter values). We then sample six parameters from the two distributions. Using these parameters, we generate snapshots of the time-series for $t = 0, 0.5, 1.0, 1.5, 2.0$ with an initial value $y(0)=0.1$ (Fig. \ref{log1}(a), Observation). Similar to the exponential growth model, the estimates through EIDGM were close to each peak (Fig.~\ref{log1}, Estimation, yellow).

\subsection{Lorenz system}
\begin{figure}[h]
    \centering
    \includegraphics[width=0.48\textwidth]{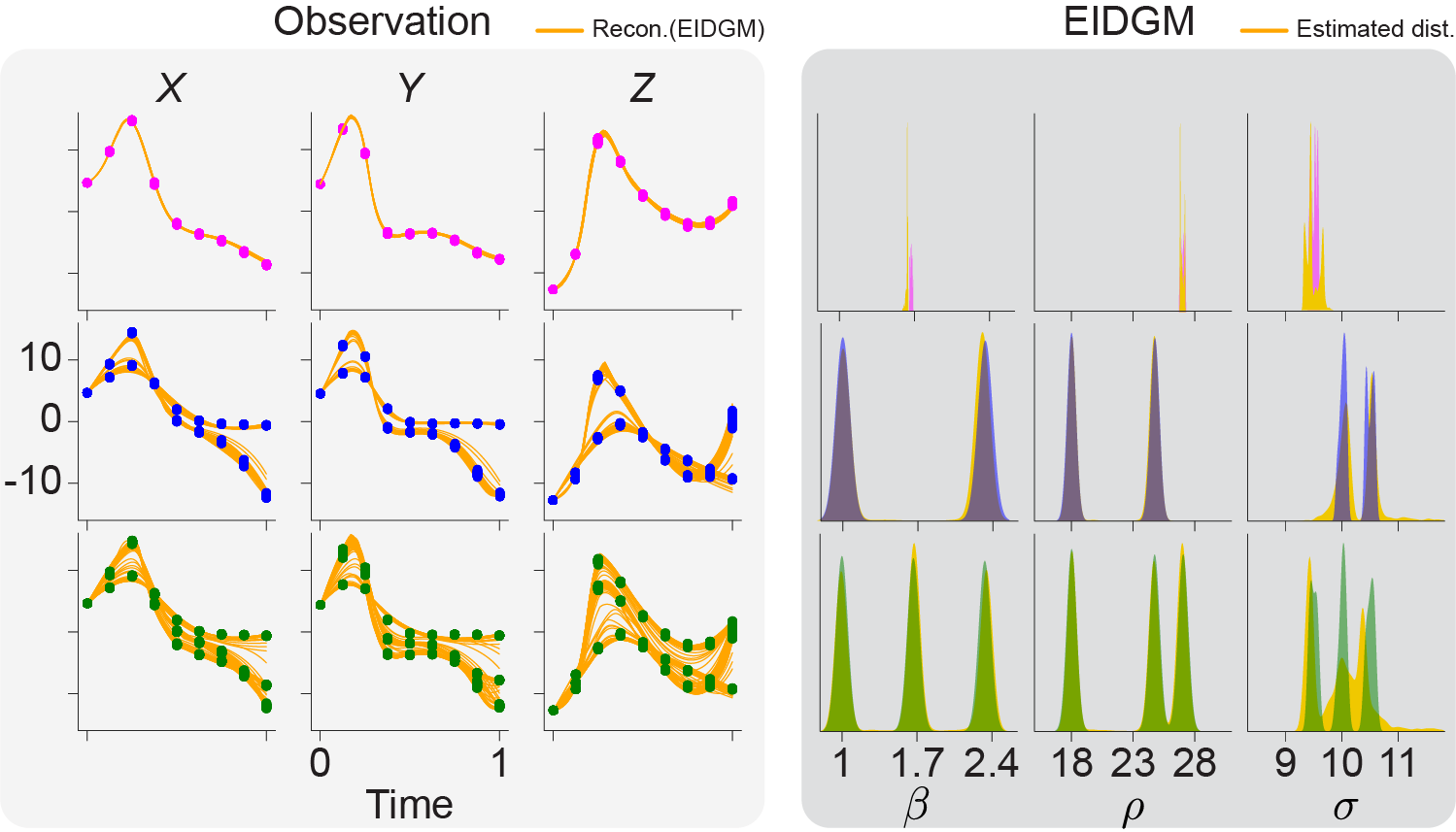}  
    \caption{{\bf Estimation of parameter distribution for the Lorenz system using EIDGM.} This model describes the temporal concentration profiles of three components: $X$, $Y$, and $Z$ (left three images in each row). Three types of parameter distributions are considered: unimodal (a), bimodal (b), and trimodal (c). For each type, we employed RCS data for three different populations (left three images) and presented the estimation results for each parameter, $\sigma$, $\rho$, and $\beta$ (rightmost image) among the six parameters included in the model.}
    \label{lorenz1}
\end{figure}

The below Lorenz system describes a simple atmospheric circulation using three key variables: the rate of convective motion in the system $X$, the temperature difference between the ascending and descending flows within the convection cell $Y$, and the deviation of the system from thermal equilibrium or the vertical temperature distribution in the convection $Z$:
\begin{align*}
\frac{dX}{dt} &= \sigma (Y-X), \nonumber\\
\frac{Y}{dt} &= X (\rho - Z) - Y, \nonumber\\
\frac{Z}{dt} &= XY - \beta Z,
\end{align*}
where Prandtl number $\sigma$ controls the ratio of fluid viscosity to thermal diffusivity, Reyleigh number $\rho$ drives convection based on temperature differences, and $\beta$ is related to the damping of convection. To evaluate the performance of EIDGM, we first generate three cases of RCS datasets (Fig. \ref{lorenz1}, left dots) based on three underlying parameter distributions (Fig. \ref{lorenz1}, right red/blue/green distributions). We then apply EIDGM to obtain parameter estimates (Fig. \ref{lorenz1}, right yellow distributions). The results demonstrate that EIDGM can accurately estimate the underlying parameter distributions. For exceptional cases, we draw the trajectories corresponding to the distributions obtained from EIDGM (Fig. \ref{lorenz1}, True-yellow lines). Surprisingly, these trajectories can penetrate the RCS dataset, suggesting potential identifiability issues in parameter estimation.

\section{Application to real-world RCS data}
\subsection{Amyloid-$\beta$ 40 and 42}
\begin{figure}[h]
    \centering
    \includegraphics[width=0.48\textwidth]{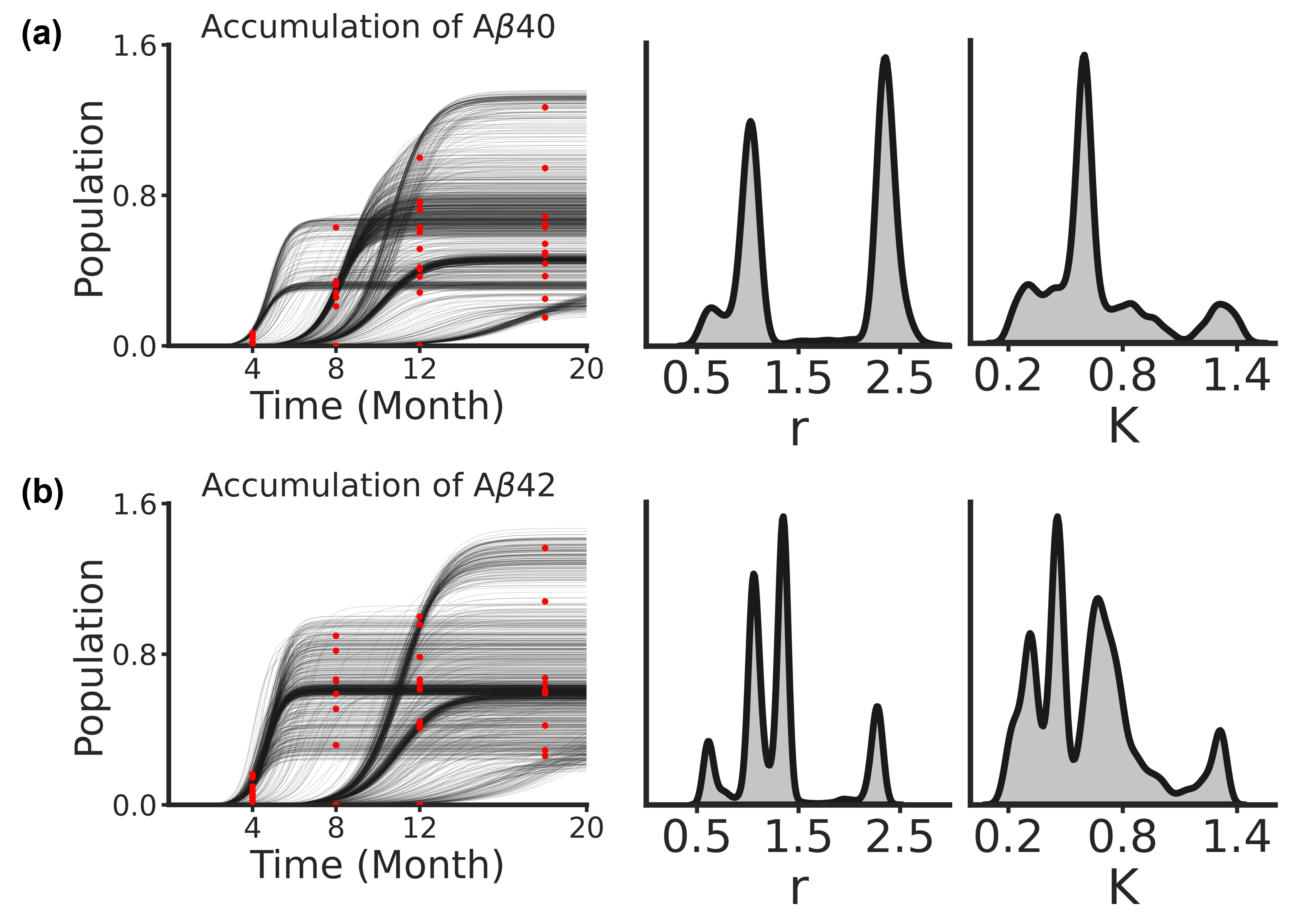}
    \caption{{\bf Estimation results for amyloid beta accumulation in real experimental datasets using a logistic model with EIDGM.} (a-b) We utilized EIDGM with two distinct datasets (left): A$\beta$40 and A$\beta$42 (left red dots). The parameter estimates revealed different patterns (right black histograms), and the corresponding trajectories accurately matched the experimental data (left translucent black lines) (a). Similar patterns were observed with the other dataset (b).}
    \label{fig:5_log_real}
\end{figure}

We estimated the growth rates and maximum population sizes for a logistic model using the amyloid-$\beta$ 40 (A$\beta$40) and amyloid-$\beta$ 42 (A$\beta$42) datasets (Fig. \ref{fig:5_log_real}(a-b), left red dots). These datasets include concentrations of A$\beta$40 and A$\beta$42 measured at four different time points (4, 8, 12, and 18 months), with each having 12 and 13 independent observations, respectively (see \cite{yada2023few, whittington2018spatiotemporal, hao2022optimal} for details). The collection of all observations is RCS data because mice with high amyloid levels were sacrificed after the observation of A$\beta$40 and A$\beta$42. We drew parameter distributions of the logistic model with the RCS dataset (Fig. \ref{fig:5_log_real}(a-b), right). 

We first validate the accuracy of parameter estimates. Unlike the simulation dataset, we cannot find the underlying distributions. Hence, we directly draw 1,000 solution trajectories of the logistic model with the parameter estimates (Fig. \ref{fig:5_log_real}(a-b), left black lines). This shows that the model solutions with estimated parameters closely match the given RCS data. Therefore, we expect that the estimated distributions are sufficiently close to the underlying parameter distributions.

Through this estimation, we found at least two patterns in the growth rate $r$ within both the A$\beta$40 and A$\beta42$ datasets. Unlike the growth rate, the maximum capacity $K$ indicates different patterns within two datasets. Specifically, A$\beta$40 converged towards a consistent maximum level of approximately 0.6 across all test subjects, whereas A$\beta$42 displayed relatively variable maximum levels that depend on the individual test subject.

\section{Conclusion}
In this paper, we introduced the EIDGM to address the challenges of parameter estimation from RCS data in dynamic system modeling. EIDGM effectively combines the strengths of deep learning in both emulating numerical solvers and estimating complex distributions by integrating HyperPINN and WGAN. The experimental results demonstrate that EIDGM significantly improves the accuracy and reliability of parameter estimation compared to traditional methods.

In EIDGM, we adopted the WGAN with gradient penalty \citep{arjovsky2017wasserstein, gulrajani2017improved}. A typical vanilla GAN framework often suffers from mode collapse and gradient vanishing issues \citep{arjovsky2017wasserstein}, which can result in critical flaws when estimating the shape of the parameter distribution. Unlike a vanilla GAN, WGAN stabilizes the training by addressing the gradient vanishing problem using the Wasserstein distance, rather than the Jensen–Shannon divergence. Thus, we applied the Wasserstein distance to improve the quality of the generated distributions.

Traditional methods, such as using mean values of trajectories or Gaussian process-based model calibration, often fail to handle the heterogeneities present in RCS data. These methods tend to produce mismatches, leading to significant incorrect parameter interpretations. EIDGM addresses these limitations by generating more precise parameter distributions, thereby providing a more accurate understanding of the dynamics in natural or engineered systems.

We validated the efficacy of EIDGM across various models, including exponential growth, logistic population models, and the Lorenz system. Our approach successfully captured the parameter distributions for these models. In the case of real-world datasets, where true parameter values are generally unknown, we used EIDGM to calculate the solution corresponding to the estimated parameters and evaluate how well it could reproduce the original RCS dataset.

Our study has several limitations. First, the accuracy of the parameters estimated using EIDGM may depend on the DE solving performance of the emulator. Although we selected hypernetworks as the emulator in this study, the choice of network type may vary depending on the specific differential equations involved. In our simulation results with a tri-modal distribution in the exponential model, DeepONet achieved better prediction performance than HyperPINN. Thus, the choice of emulator may depend entirely on the type of equation. Second, we did not introduce any noise into the RCS dataset during the simulation, as there was no clear way to distinguish whether the variations in parameter estimates were due to noise or inherent heterogeneity in the data. This issue, known as the identifiability problem, emerged due to the complexities in managing RCS data. Therefore, more comprehensive analyses of RCS data are required in future research.

Applying EIDGM to real-world datasets further showcased its ability to handle diverse shapes of parameter distributions, highlighting its broad application potential in fields, such as economics, political science, and biology. This demonstrates that EIDGM's ability to provide precise parameter estimations opens up new opportunities for scientific research in understanding complex dynamical systems with limited data availability.

\section{Acknowledgement}
Hyung Ju Hwang was supported by the National Research Foundation of Korea grant funded by the Korea government (RS-2023-00219980 and RS-2022-00165268). Hyeontae Jo was supported by the National Research Foundation of Korea (RS-2024-00357912) and Korea University grant K2411031. Sung Woong Cho 
was supported by the National Research Foundation of Korea (NRF) grant funded by the Ministry of Science and ICT (KR) (RS-2024-00462755)

\bibliographystyle{plain}
\bibliography{bib_EIDGM.bib}
\newpage
\section{Biography Section}
\textbf{Hyunwoo Cho} received the B.S. degree in mathematics from POSTECH, in 2019, where he is currently pursuing the Ph.D. degree in math ematics. His research interests include the applications of machine learning and physics-informed machine learning.

\textbf{Sung Woong Cho} received his B.S. degree in mathematics and Ph.D. degree from Postech in 2023. He is currently a postdoctoral researcher at the Stochastic analysis and application Research Center (SaaRC) located in the Korea Advanced Institute for Science and Technology (KAIST). His research interest is mainly focused on deep learning and partial differential equations.

\textbf{Hyeontae Jo} received the Ph.D. degree in mathematics from POSTECH, Pohang, South Korea, in 2020. He is currently an assistant professor of applied mathematics from Korea University, South Korea, in 2024. His research interests include mathematical modeling and artificial intelligence, with a focus on scientific discovery.

\textbf{Hyung Ju Hwang} received her Ph.D. degree in mathematics from Brown University, in 2002. She is currently a chaired professor in the Department of Mathematics, POSTECH. Prior to this, she was a Research Assistant rofessor with Duke University, from 2003 to 2005, and a Postdoctoral Researcher with Max-Planck Institute, Leipzig, from 2002 to 2003. She has published more than 80 scientific articles in the fields of applied mathematics and mathematical AI. Her research interests include optimization, deep learning, applied mathematics, partial differential equations, and data analysis in applied fields.

\clearpage
\newpage

\appendix

\section{Supplemental materials}

\subsection{Mathematical analysis}
This section presents a mathematical analysis of the EIDGM and it can accurately estimate parameter distributions that fit the model (Eq.~(\ref{equation:ODE})) to RCS data. There are three steps as follows: 1) We first establish that HyperPINN can accurately generate a solution to Eq.~(\ref{equation:ODE}) for a given parameter set $\mathbf{p}$ by training with both data and physics loss functions, as defined in Eq.~(\ref{equation:loss_phy}). As it cannot be realized on a computational device, we use discretized versions of the data and physics loss functions. 2) We next show that the discretization does not significantly affect the accuracy of the HyperPINN framework up to a specific error bound. 3) Finally, we demonstrate that the discretized version of the physics loss function can be minimized with sufficiently many weights and biases in HyperPINN. In summary, HyperPINN can effectively operate as a DE emulator, calculating the solutions for a given parameter set $\mathbf{p}$ by minimizing the discretized losses.

\subsubsection{Training HyperPINN by minimizing both data and physics losses}
Raissi \textit{et al.} introduced the original physics loss \cite{raissi2019physics}, $\ell(nn(t), \theta_{nn})$, which measures how well an artificial neural network $nn(t)$ with weights and biases $\theta_{nn}$ satisfies the given dynamical system of Eq.~(\ref{equation:ODE}):
\begin{equation}
    \ell(nn(t), \theta_{nn})  = \left(\frac{d}{dt}nn(t;\theta_{nn})-f[nn(t;\theta_{nn}), \mathbf{p}, t]\right)^{2}\label{l_phys_general_l}.
\end{equation}
For a given fixed $\mathbf{p}$, previous studies have shown that neural networks $nn(t)=nn(t; \theta_{nn})$ closely approximate the solution $\mathbf{y}(t;\mathbf{p})$ of Eq.~(\ref{equation:ODE}) when $\ell(nn(t), \theta_{nn})$ becomes small over every $t$ \cite{sirignano2018dgm, raissi2019physics, jo2019deep, cho2023monotone, li2024physics}. Specifically, we refer to the following theorem in \cite{hirsch2013differential}, which can be derived by Gr\"onwall's and H\"older's inequalities.
\begin{theorem}\label{appendix:Theorem_stability}
Suppose that \( f[\mathbf{y}(t), \mathbf{p}, t] \) is Lipschitz continuous in \( \mathbf{y} \) with Lipschitz constant \( L > 0 \). Assume that the the neural network $nn(t)$ satisfies \( \left|\mathbf{y}\left(t_{1}; \mathbf{p}\right)-nn\left(t_{1}; \theta_{nn}\right)\right| \leq \delta \) for some \( \delta \geq 0 \), where $\mathbf{y}(t; \mathbf{p})$ is the solution of Eq.~(\ref{equation:ODE}). Then, the following inequality holds:
\begin{multline*}
\left|\mathbf{y}(t;\mathbf{p}) - nn(t; \theta_{nn})\right| \\
\leq \Bigg(\delta + \Big[(t - t_{1}) 
\int_{t_{1}}^{t_{T}} \ell(nn(t), \theta_{nn}) \, dt \Big]^{\frac{1}{2}} \Bigg) \\
\times e^{L\left(t - t_{1}\right)}, \quad 
\text{for} \quad \forall t \geq t_{1}.
\end{multline*}
\end{theorem}

As we use hyperPINN to calculate the solutions for various sets of parameters $\mathbf{p}$ simultaneously, the original physics loss can be modified. More specifically, weights and biases $\theta_m(\mathbf{p})$ in the main network $m(t)$ in HyperPINN are determined by the output of the other networks, i.e., $\theta_m(\mathbf{p})=h(\mathbf{p}, \theta_h)$. Hence, we take an expectation $\mathbb{E}$ over the probability density function of the parameter $\mathbf{p}$, $\mathcal{D}$, on the physics loss:
\begin{align}
    L_{\text{physics}}(\theta_h)& = \mathbb{E}_{\mathbf{p}\sim \mathcal{D}} \ell(m(t),  \theta_m(\mathbf{p})).\label{l_phys_general}
\end{align}
By minimizing this loss function, we expect that the solutions from the main network with a parameter $\mathbf{p}$, sampled from $\mathcal{D}$, can accurately satisfy Eq.~(\ref{equation:ODE}). To verify this, we first obtain the lower bound on the probability that HyperPINN prediction for given parameters $\mathbf{p}$ has a small value of $\ell(m(t), \theta_{m}(\mathbf{p}))$: 
\begin{align*}
    P\bigg(&\frac{1}{T_{col}} \sum_{i=1}^{T_{col}} 
    \ell(m(t_{i}^{c}), \theta_{m}(\mathbf{p})) \leq \varepsilon\bigg) \\
    &\geq \prod_{i=1}^{T_{col}} 
    P\big(\ell(m(t_{i}^{c}); \theta_{m}(\mathbf{p})) \leq \varepsilon\big) \\
    &\geq \prod_{i=1}^{T_{col}} 
    \bigg(1 - \frac{\mathbb{E}_{\mathbf{p}\sim \mathcal{D}} 
    \ell(m(t_i^{c}), \theta_{m}(\mathbf{p}))}{\varepsilon}\bigg) \\
    &\geq 1 - \sum_{i=1}^{T_{col}} 
    \frac{\mathbb{E}_{\mathbf{p}\sim \mathcal{D}} 
    \ell(m(t_i^{c}), \theta_{m}(\mathbf{p}))}{\varepsilon} \\
    &= 1 - \frac{T_{col}}{\varepsilon} L_{\text{physics}}(\theta_h).
\end{align*}
where the second inequality is obtained using the standard Markov's inequality. Therefore, for a given parameter $\mathbf{p}$ sampled from $\mathcal{D}$, the original physics loss $\ell(m(t), \theta_{nn})$ in (\ref{l_phys_general_l}) can be reduced within an error $\varepsilon$ if $L_{\text{physics}}$ is sufficiently small. Consequently by Theorem \ref{appendix:Theorem_stability}, a small value of $\ell(m(t), \theta_{nn})$ implies that the main network $m(t; \theta_{m}(\mathbf{p}))$ is close to the solution $\mathbf{y}(t, \mathbf{p})$ of Eq.~(\ref{equation:ODE}).

\subsubsection{Difference of the two physics loss functions}
In practice, we cannot calculate the above definite integral and expectation due to the lack of computational resources. Thus, we alternatively use the discretized version of $L_{\text{physics}}$ defined in Eq.~(\ref{equation:loss_phy(disc)}) as follows:
\begin{equation*}
L_{\text{physics(disc)}}(\theta_h) = \frac{1}{T_{col}}\sum_{i=1}^{T_{col}}\frac{1}{N_{p}}\sum_{j=1}^{N_{p}} \ell(m(t_{i}), \theta_{m}(\mathbf{p}_{j})), \nonumber
\end{equation*}
where $N_{p}$ denotes the number of trajectories used in the discretized physics-informed loss $L_{\text{physics(disc)}}(\theta_h)$.
The set of $N_p$ parameters $\{\mathbf{p}_{j} \}_{j=1}^{N_p}$ are sampled from the probability distribution $\mathcal{D}$. Sequentially, a set of solutions $\{\mathbf{y}(t;\mathbf{p}_{j})\}_{j=1}^{N_p}$ is obtained through the DE solver (e.g., LSODA in \textit{Scipy} package). 

Despite the discretization, we first show that the value of $L_{\text{physics(disc)}}(\theta_h)$ can be close to $L_{\text{physics}}(\theta_h)$ up to specific error bounds. To show this, we briefly introduce some definitions required for the proof. 
Let $\mathcal{M}$ be a class of main networks, where each main network is associated with a hypernetwork. Hypernetworks in each main network, $h$, consist of fully connected neural networks with different weights and biases of the $i$-th layer, $[W, b]=\{W^i \in \mathbb{R}^{g_{i+1}\times g_{i}}, b^i\in \mathbb{R}^{g_{i+1}}\}_{i=1}^{n}$, and map $\mathbf{p}$ to the weights and biases of the main networks. For convenience, we denote a single element of $\mathcal{M}$ as $m(t)=m(t; h(\mathbf{p}, [W, b]))=m(t; h(\mathbf{p}; \theta_{h}))$, where $\theta_{h}$ refers the weights $W$ and biases $b$.      

Using $\mathcal{M}$, we first prove that there exists a subset of $\mathcal{M}$ with a finite number of elements such that every element in $\mathcal{M}$ is close to at least one element in this subset up to distance $\varepsilon>0$. The distance $d_{D}$ between two elements $m(t)$ and $\overline{m}(t)$ in $\mathcal{M}$ with hypernetworks $h(\cdot, \theta_{h})$ and $\bar{h}(\cdot, \theta_{\bar{h}})$ is defined by 
\begin{multline*}
    d_D(m, \overline{m}) := \frac{1}{T_{col}} \sum_{i=1}^{T_{col}} 
    \mathbb{E}_{\mathbf{p} \sim D} \big|\ell(m(t_i^{c}), h(\mathbf{p}; \theta_h)) \\
    - \ell(m(t_i^{c}), \bar{h}(\mathbf{p}; \theta_{\bar{h}}))\big|.
\end{multline*}

The subset is called an $\varepsilon$-cover of \((\mathcal{M}, d_\mathcal{D})\). The rigorous definition of an $\varepsilon$-cover is as follows:
\begin{definition}\label{epsilon_cover}
A set \( \{m_1, m_2, \cdots, m_c\} \) in \(\mathcal{M} \) is called an $\varepsilon$-cover of \((\mathcal{M}, d_\mathcal{D})\) if for every main network in $\overline{m}\in\mathcal{M}$, we can always find an index $i\in\{1,2,...,c\}$ such that \(d_{\mathcal{D}}(m_{i}, \overline{m})<\varepsilon \). 
\end{definition}
In Definition \ref{epsilon_cover}, we denote an integer $\mathcal{N}(\varepsilon, \mathcal{M}, d_\mathcal{D})$ as the smallest number of $\varepsilon$-covers of \((\mathcal{M}, d_{\mathcal{D}})\). Sequentially, we define the covering number that indicates the number of candidates in the class as follows: 
\begin{equation}\label{covering_number}
    \mathcal{C}(\varepsilon, \mathcal{M}):=\sup_{\mathcal{D}\in\mathbb{D}} \mathcal{N}(\varepsilon, \mathcal{M}, d_{\mathcal{D}}),
\end{equation}
where $\mathbb{D}$ denotes the set of probability distributions of parameter $p$ that share the same compact support. In a previous study, when time interval $[t_{1}, t_{T}]$ is just a point (i.e., $t_{1}=t_{T}$), Baxter \textit{et al.} showed that the absolute difference between $L_{\text{physics}}(\theta_{h})$ and $L_{\text{physics(disc)}}(\theta_{h})$ has an upper bound, depending only on the covering number \cite{baxter2000model}.
\begin{theorem}\label{appendix:theorem_interpolation(one timepoint)}
(Theorem 3, \cite{baxter2000model}) Given $\varepsilon>0$ and $1>\delta>0$, suppose that $T_{col}=1$ and the number of trajectories in Eq.~(\ref{equation:loss_phy}) for training, $N_{p}$, satisfies the following inequality:
\begin{equation}\nonumber
    N_{p}\geq \max \left\{\frac{64}{\varepsilon^2}\log\frac{4\mathcal{C}(\frac{\varepsilon}{16}, \mathcal{M})}{\delta}, \frac{16}{\varepsilon^{2}} \right\}.
\end{equation}
Then, the following inequality holds with probability at least $1-\delta$:
\begin{equation}\label{loss_discretization}
    |L_{\text{physics(disc)}}(\theta_h)- L_{\text{physics}}(\theta_{h})| \leq \varepsilon.
\end{equation}

\end{theorem}

Note that Theorem \ref{appendix:theorem_interpolation(one timepoint)} is available only when $T_{col}=1$. In general, the number of collocation time points $T_{col}$ can be recorded multiple times. Hence, we extend the same results in the case when $T_{col}>1$ through the following corollary: 
\begin{corollary}\label{appendix:collorary_minimum_sample}
    If $N_{p}$ satisfies
\begin{equation}\nonumber
    N_{p}\geq \max \left\{\frac{64}{\varepsilon^2}\log\frac{4\mathcal{C}(\frac{\varepsilon}{16}, \mathcal{M})}{\delta/T_{col}}, \frac{16}{\varepsilon^{2}}\right\},
\end{equation}
then, the same inequality of Eq.~(\ref{loss_discretization}) in Theorem \ref{appendix:theorem_interpolation(one timepoint)} also holds.
\end{corollary}

\begin{proof}
    Suppose that we derive the $N_{p}$ trajectories as above. Then, for any $m\in \mathcal{M}$, we can guarantee that $P(A_{i})\geq 1-\delta/T_{col}$ for each $i$ by using Theorem \ref{appendix:theorem_interpolation(one timepoint)}, where   
    \begin{multline*}
    A_{i} = \Bigg\{ \Bigg| \frac{1}{N_{p}} \sum_{j=1}^{N_{p}} 
    \ell(m(t^{c}_{i}), \theta_{j}(\mathbf{p}_j)) \\
    - \int_{\mathcal{P}} \ell(m(t^{c}_{i}), \theta(\mathbf{p})) \, dD(\mathbf{p}) 
    \Bigg| \leq \epsilon \Bigg\}.
    \end{multline*}
    Because \( P(\bigcap_{i=1}^{T_{col}}A_{i}) \geq 1-\sum_{i=1}^{T_{col}} P(A_i ^C)\geq 1-T_{col} \cdot (\delta/T_{col}) = 1-\delta \), where $A_i^C$ is the complement of $A_i$, we can derive the corollary by using the triangle inequality.
    \begin{align*}
    &\Bigg| \frac{1}{T_{col}} \sum_{i=1}^{T_{col}} \frac{1}{N_{p}} 
    \sum_{j=1}^{N_{p}} \ell(m(t^{c}_{i}), \theta_{j}(\mathbf{p}_j)) \\
    &\quad - \frac{1}{T_{col}} \sum_{i=1}^{T_{col}} 
    \int_{\mathcal{P}} \ell(m(t^{c}_{i}), \theta(\mathbf{p})) \, dD(\mathbf{p}) \Bigg| \\
    &\leq \frac{1}{T_{col}} \sum_{i=1}^{T_{col}} \Bigg| \frac{1}{N_{p}} 
    \sum_{j=1}^{N_{p}} \ell(m(t_i^{c}), \theta_{j}(\mathbf{p}_j)) \\
    &\quad - \int_{\mathcal{P}} \ell(m(t_{i}^{c}), \theta(\mathbf{p})) \, dD(\mathbf{p}) \Bigg|.
    \end{align*}
\end{proof}
According to Theorem \ref{appendix:theorem_interpolation(one timepoint)} and Corollary \ref{appendix:collorary_minimum_sample}, the number of parameters $N_p$ depends on the covering number $\mathcal{C}$. However, $\mathcal{C}$ is generally not finite when the set of probability distributions of the parameter $p$, $\mathbb{D}$, does not share a common compact support. To address this, we added an assumption in the definition of the covering number (\ref{covering_number}) that requires a shared compact support. This assumption is practically reasonable because, in real-world scenarios, parameter estimation in differential equations typically involves bounded parameter ranges that are verified through experiments. Moreover, the set of all main networks $\mathcal{M}$ yields the same issue; hence, we assume that the weights and biases in main networks are bounded. 

Under this assumption, we demonstrate that \(\mathcal{C}(\frac{\varepsilon}{16}, \mathcal{N}_{\ell})\) is indeed finite when parameter distributions in $\mathbb{D}$ share the same compact support. Without loss of generality, the compact support is contained in the finite interval $[p_{min}, p_{max}]^{n_p}$, where $n_{p}$ is the dimension of the parameters $\mathbf{p}$. For this proof, we first verify that our hypernetwork and loss function exhibit Lipschitz continuity. We also assume that all the components in the weights and biases of neural networks are contained in $[-R, R]$. With this assumption, we can obtain the following lemma:     
\begin{lemma} \label{appendix:lemma_lipschitz}
    Let $h(\mathbf{p}; \theta_h):=W^{k}(\sigma(W^{k-1}\cdots\sigma(W^{1}\mathbf{p}+ b^{1})+b^{k-1})+b^{k}$, where \(W^{i}\in [-R, R]^{g_{i+1}\times g_{i}}\) and $b^{k}\in [-R, R]^{g_{i}}$. $\sigma$ is a Lipschitz continuous activation function with Lipschitz constant $L_{\sigma}$. Suppose that $f[m(t;\theta_m), \mathbf{p}, t]$ has some constant $L_{f}$ such that the following holds:
    \begin{equation*}
    \|f[m(t;\theta_m), \mathbf{p}, t] -f[m(t;\theta'_m), \mathbf{p}, t]\|_{1}\leq L_{f}\|\theta_{m}-\theta'_m\|_{1} 
    \end{equation*}
    Then, there exist constants \(L_{p}, L_{h}, L_{\ell}\) depending on \( L_{\sigma}, R, \{g_{i}\}_{i=1}^{k+1} \) such that the following holds:
    \begin{align*}
    &\|h(\mathbf{p}_{1}; \theta_{h}) - h(\mathbf{p}_{2}; \theta_{h})\|_{1} 
    \leq L_{p} \|\mathbf{p}_{1} - \mathbf{p}_{2}\|_{1}, \\
    &\|h(\mathbf{p}; \theta_{h}^{1}) - h(\mathbf{p}; \theta_{h}^{2})\|_{1} 
    \leq L_{h} \|\theta_{h}^{1} - \theta_{h}^{2}\|_{1}, \\
    &\|\ell(m(t), h(\mathbf{p}_{1}; \theta_{h}^{1})) - \ell(m(t), h(\mathbf{p}_{2}; \theta_{h}^{2}))\|_{1} \\
    &\quad\quad\quad\quad\quad\quad \leq L_{\ell} 
    \|h(\mathbf{p}_{1}; \theta_{h}^{1}) - h(\mathbf{p}_{2}; \theta_{h}^{2})\|_{1}.
    \end{align*}
\end{lemma}
\begin{proof}
    We derive the first Lipschitz continuity using the triangle inequality and the following inductive step: 
    \begin{align*}
    &\|h(\mathbf{p}_1;\theta_{h}) - h(\mathbf{p}_2;\theta_{h})\| \\
    &= \Big\|W^k\big(\sigma(W^{k-1}\cdots\sigma(W^1\mathbf{p}_1 + b^1) + b^{k-1}) \\
    &\quad - \sigma(W^{k-1}\cdots\sigma(W^1\mathbf{p}_2 + b^1) + b^{k-1})\big)\Big\| \\
    &\leq \|W^k\| \cdot L_{\sigma} \cdot 
    \Big\| (W^{k-1}\cdots\sigma(W^1\mathbf{p}_1 + b^1) + b^{k-1}) \\
    &\quad - (W^{k-1}\cdots \sigma(W^1\mathbf{p}_2 + b^1) + b^{k-1}) \Big\|_{1} \\
    &\leq \cdots \\
    &\leq L_{\sigma}^{k-1} \|W^k\| \cdots \|W^2\| 
    \Big\|(W^1\mathbf{p}_1 + b^1) - (W^1\mathbf{p}_2 + b^1)\Big\| \\
    &\leq L_{\sigma}^{k-1} \|W^k\| \cdots \|W^1\| \|\mathbf{p}_1 - \mathbf{p}_2\| \\
    &\leq L_{\sigma}^{k-1}(Rg_{k+1}g_k)\cdots(Rg_2g_1)\|\mathbf{p}_1 - \mathbf{p}_2\| \\
    &= L_{\sigma}^{k-1} R^k \frac{(g_1 \cdots g_{k+1})^2}{g_1 g_{k+1}} \|\mathbf{p}_1 - \mathbf{p}_2\|.
    \end{align*}

    To compute the left-hand side of the second inequality, we define the hypernetwork $h(\mathbf{p};\theta_h')$ with weights and biases different from those of $h(\mathbf{p}, \theta_{h})$ (but same depth and width) as follows: 

    \begin{align*}
        h(\mathbf{p};\theta_{h}^{'})= V^{k} \sigma (V^{k-1}\cdots \sigma(V^{1}(\mathbf{p})+c^{1})+c^{k-1})+c^{k}.
    \end{align*}
    Using the triangle inequality, we derive the following relationship between the $k-$th and $(k-1)-$th layers:
    \begin{align*}
    &\|h(\mathbf{p};\theta_{h}) - h(\mathbf{p};\theta_{h}')\| \\
    &\leq \Big\| W^{k} \sigma\big(W^{k-1} \cdots \sigma(W^{1}(\mathbf{p}) + b^{1}) 
    + b^{k-1}\big) \\
    &\quad - V^{k} \sigma\big(V^{k-1} \cdots \sigma(V^{1}(\mathbf{p}) + c^{1}) 
    + c^{k-1}\big) \Big\|  + \|b^{k} - c^{k}\| \\
    &\leq \Big\| W^{k} \big(\sigma(W^{k-1} \cdots \sigma(W^{1}(\mathbf{p}) + b^{1}) 
    + b^{k-1}) \\
    &\quad - \sigma(V^{k-1} \cdots \sigma(V^{1}(\mathbf{p}) + c^{1}) + c^{k-1}) \big) \Big\| \\
    &\quad + \|W^{k} - V^{k}\| \cdot \sigma\big(V^{k-1} \cdots \sigma(V^{1}(\mathbf{p}) 
    + c^{1}) + c^{k-1}\big) \\
    &\quad + \|b^{k} - c^{k}\| \\
    &\leq \|W^{k}\| \cdot L_{\sigma} \cdot \Big\| W^{k-1} \cdots \sigma(W^{1}(\mathbf{p}) 
    + b^{1}) + b^{k-1} \\
    &\quad - \big(V^{k-1} \cdots \sigma(V^{1}(\mathbf{p}) + c^{1}) + c^{k-1}\big) \Big\| \\
    &\quad + Rg_{k} \|\theta_{h} - \theta_{h}'\| + \|\theta_{h} - \theta_{h}'\|.
    \end{align*}
    Finally, we observe that the first layer should be Lipschitz continuous:
    \begin{align*}
        &\|(W^{1}(\mathbf{p})+b^{1})-(V^{1}(\mathbf{p})+c^{1})\|
        \\& \quad\quad\leq \|W^{1}-V^{1}\| \|\mathbf{p}\|+ \|b^{1}-c^{1}\|\\&\quad\quad\leq\|\theta_{h}-\theta_{h}^{'}\| Rg_{1} +\|\theta_h-\theta_{h}^{'}\|.
    \end{align*}
    Therefore, we derive the second inequality using the inductive step.

    For the last inequality, note that $\theta_{m}(\mathbf{p})=h(\mathbf{p};\theta_{h})$ is bounded by the first inequality. This implies that $m(t_{i};\theta_m)$ and $f(m(t_{i}; \theta_m))$ should be bounded because $f$ is Lipschitz continuous. Letting $m(t;\theta_{m})=U^{k}\sigma(U^{k-1}\cdots \sigma(U^{1}(t)+b^{1})+b^{k-1})+b^{k}$, its derivative can be obtained by 
    \begin{align*}
        (\Pi_{i=2}^{k}(U^{i}\sigma'(U^{i-1}\cdots\sigma(U^{1}(t)+a^{1})+a^{i-1})))U^{1}, 
    \end{align*}
    where $\sigma'(x)$ denotes a diagonal $g_{i}\times g_{i}$ matrix with its diagonal component equal to the derivative of $\sigma(x)$ when $x\in \mathbb{R}^{g_{i}}$. By the first inequality proved above, $U^{i}\sigma'(U^{i-1}\cdots\sigma(U^{1}(t)+a^{1})+a^{i-1})$ are bounded and Lipschitz. Because the product of bounded Lipschitz functions is also bounded, we conclude that \(dm(t_{i};\theta_{m})/dt  \) is bounded and Lipschitz. Finally, in the same way, \( \sum_{i=1}^{T} (\frac{d}{dt}m(t_i;\theta_{m})-f(m(t_{i};\theta_{m}))^{2}) \) is Lipschitz with its constant depending on $R, L_{\sigma}, \{g_{i}\}_{i=1}^{k+1}$
\end{proof}
From  Lemma \ref{appendix:lemma_lipschitz}, we can derive the finite upper bound of the covering number as follows:
\begin{lemma} \label{appendix:lemma_covering}
    Suppose that the same assumption holds for the parameter in Lemma \ref{appendix:lemma_lipschitz}. For a given $\varepsilon>0$, the following inequality holds:
    \begin{align*}
        &\mathcal{C}\left(\frac{\varepsilon}{16}, \mathcal{N}_{\ell}\right)
        \\&\quad\leq \left(\frac{\epsilon}{16L_{\ell}L_h\sum_{i=1}^{k}g_{i}g_{i+1}+g_{i+1}}\right)^{\sum_{i=1}^{k}g_{i}g_{i+1}+g_{i+1}}
    \end{align*}
\end{lemma}
\begin{proof}
We first calculate the distance between two classes of neural networks. Assuming that the main network and hypernetwork architectures are identical, with differences only in the weights and biases of the hypernetwork, the distance can be computed as follows:
\begin{align*}
&d_D\left( m(t;h(\cdot;\theta_h^{1})), m(t;h(\cdot;\theta_{h}^{2})\right)\\&= \frac{1}{T_{col}}\sum_{i=1}^{T_{col}}\int_{\mathcal{P}} \ell(m(t_{i}^{c}), h(\mathbf{p}; \theta_{h}^{1})) - \ell(m(t_{i}^{c}), h(\mathbf{p}; \theta_{h}^{2}) )dD(\mathbf{p})
\end{align*}

Using Lemma \ref{appendix:lemma_lipschitz}, we obtain the following inequality:

\begin{align*}
&\int_{\mathcal{P}} \ell(m(t), h(\mathbf{p}; \theta_{h}^{1})) - \ell(m(t), h(\mathbf{p}; \theta_{h}^{2}) )dD(\mathbf{p})  \\&\leq 
\int_{\mathcal{P}} L_{\ell}\|m(t;h(\cdot;\theta_h^{1})) - m(t;h(\cdot;\theta_h^{2})) \|dD(\mathbf{p})
\\ &\leq \int_{\mathcal{P}}L_{\ell}L_{h}\|\theta_h^{1} - \theta_h^{2}\| dD(\mathbf{P})= L_{\ell}L_{h}\|\theta_h^{1} - \theta_h^{2}\|
\end{align*}

Now, if we select a covering of the weight and bias space such that each \( \theta_h^{1} \) has a close weight \( \theta_h^{2} \) at a distance of \( \epsilon/16 L_{\ell}L_{h} \), then we get an \( \frac{\epsilon}{16} \)-covering in the \(d_D\left( m(t;h(\cdot;\theta_h^{1}), m(t;h(\cdot;\theta_h^{2})\right) \) metric. Because the total number of weights and biases is $\sum_{i=1}^{k}(g_{i}g_{i+1}+g_{i+1})$, $\|\theta_h^{1} -\theta_h^{2}\|_{1}\leq \epsilon/16L_{\ell}L_h$ if each component of $\theta_h^{1}$ and $\theta_h^{2}$ lies within $\epsilon/16L_{\ell}L_h\sum_{i=1}^{k}(g_{i}g_{i+1}+g_{i+1})$. Given that all weights and biases lie within $[-R, R]$, we can conclude that the desired result holds.

\end{proof}

By combining the results of Corollary \ref{appendix:collorary_minimum_sample} and Lemma \ref{appendix:lemma_covering}, we derive Theorem \ref{appendix:theorem_interpolation}. This theorem implies that by appropriately sampling the parameters $\{p_{j}\}_{j=1}^{N_{p}}$ and minimizing the discretized physics loss function $L_{\text{physics(disc)}}(\theta_h)$, we can effectively minimize the physics loss function $L_{\text{physics}}$. 
\begin{theorem}\label{appendix:theorem_interpolation}
    Suppose that the same assumption holds for the parameter in Lemma \ref{appendix:lemma_lipschitz}. The number $N_{p}$ of sampled trajectories for training satisfies the following:
    \begin{align*}
    N_{p} &\geq \max \Bigg\{ 
    \frac{64}{\varepsilon^2} \sum_{i=1}^{k} 
    \big(g_{i}g_{i+1} + g_{i+1}\big)\times \\
    &\quad\quad\log\bigg(\frac{\varepsilon}{4LL_{h}\delta 
    \sum_{i=1}^{k} \big(g_{i}g_{i+1} + g_{i+1}\big) / T_{col}}\bigg), \frac{16}{\varepsilon^{2}} 
    \Bigg\}.
    \end{align*}

Then, the following inequality holds with probability at least \(1-\delta \):
    \[|L_{\text{physics(disc)}}(\theta_h) - L_{\text{physics}}(\theta_{h})|\leq \varepsilon   \]
\end{theorem}

\subsubsection{Minimizing the discretized loss function using synthetic RCS data}

The previous section guarantees that the physics loss can be discretized up to a small error when we train it with a sufficient number of parameters. In this section, we demonstrate that hyperPINN can effectively minimize $L_{\text{physics(disc)}}$ by applying the universal approximation theorem for neural networks. According to this theorem, a neural network can approximate any differentiable function given a sufficient number of weights and biases.

\begin{theorem}[Theorem 2.1 in \cite{li1996simultaneous}]\label{appendix:theorem_universal}
     Let $K$ be a compact subset of $\mathbb{R}^d$. Suppose that all of the first partial derivatives of $f$ lie in $C^m(\Omega)$ for some open set $\Omega$ containing $K$. If the activation function $\sigma$ belongs to $C^1(\mathbb{R})$, then for any $\varepsilon>0$, there exists a 2-layer fully connected neural network $u_{nn}(x)=\sum_{i=1}^{h}c_i \sigma(w_i x+b_i)$ such that 
    \begin{equation}
        ||D^\alpha (u_{nn}) - D^\alpha (f)||_{L^{\infty}(K)}<\varepsilon, \forall \alpha \in Z_{+}^d  \text{ with } |\alpha|\le m. \nonumber
    \end{equation}
\end{theorem}

By applying the above theorem, we derive Proposition \ref{appendix:proposition_universal}, which guarantees the existence of hyperPINN with an arbitrarily low value of the loss function $L_{\text{physics}}$. Consequently, Theorem \ref{appendix:theorem_interpolation}, in conjunction with the proposition below, implies that with sufficiently large sample size, hyperPINN can ensure a low value of $L_{\text{physics}}$ for all parameters $\mathbf{p}$.

\begin{proposition} \label{appendix:proposition_universal}
    Suppose that Eq.~(\ref{equation:ODE}) has a unique solution $\mathbf{y}_{j}$ for each parameter $\{\mathbf{p}_{j}\}_{j=1}^{N_{p}}$. Further, assume that the conditions of Lemma \ref{appendix:lemma_lipschitz} hold and that $R$ is sufficiently large. Then, for any given $\varepsilon>0$, there exists $\theta_h$ such that the following inequality holds: 
        \[L_{\text{physics}(disc)}(\theta_h)\leq\varepsilon.\]
\end{proposition}
\begin{proof}
    By Theorem \ref{appendix:theorem_universal}, for a given $\epsilon>0$ and parameters $\{\mathbf{p}_{j}\}_{j=1}^{N_{p}}$, there exists a fully connected neural network $m_{j}$ with weights and biases $\theta_{m_{j}}$ such that 
    \begin{multline*}
    |m(t^c_{i}, \theta_{m_{j}}) - \mathbf{y}_{j}(t^c_{i})| + \left|\frac{dm}{dt}(t^{c}_{i}, \theta_{m_{j}}) - \frac{d\mathbf{y}_{j}}{dt}(t^{c}_{i})\right| \\
    \leq \frac{\sqrt{\varepsilon}}{\sqrt{2 + 2L_{f}^{2}}},
    \end{multline*}
    where $L_{f}$ is the Lipschitz constant of $f$, as defined in Lemma \ref{appendix:lemma_lipschitz}.
    
    Consider a function $\bar{h}:[p_{min}, p_{max}]^{n_{p}}\rightarrow \mathbb{R}^{|\theta_{m_{j}}|}$ in which $|\theta_{m_{j}}|$ denotes the total number of weights and biases in $m_{j}$. We can assume that $|\theta_{m_{j}}|$ is identical for all $j$ by considering the largest neural network among $\{m_{j}\}_{j}$. For any $j$ with $|\theta_{m_{j}}|$ smaller than that of the largest network, we may extend it to a fully connected neural network by adding zero-valued weights and biases.     
    
    Set $\bar{h}(\mathbf{p}_{j})=\theta_{m_{j}}$. Then, we can find the neural network $h$ with weights and biases $\theta_h$ such that
    \begin{align*}
        |h(\mathbf{p}_{j}, \theta_{h})-\bar{h}(\mathbf{p}_{j})|\leq \frac{\varepsilon}{2L_{\ell}}, \forall j\in \{1, \cdots, N_{p}\},
    \end{align*}
    where $L_{\ell}$ is the Lipschitz constant from Lemma \ref{appendix:lemma_lipschitz}. 
    
    Using the triangle inequality and the Lipschitz continuity of $f$, we derive the following inequality: 
    \begin{align*}
    &|\ell(m_{j}(t_{i}^{c}), \theta_{m_{j}}(\mathbf{p}_{j}))| \\
    &= \Bigg(\frac{d}{dt} m_{j}(t_{i}^{c}; \theta_{m_{j}}) 
    - f[m(t_{i}^{c}; \theta_{m}), \mathbf{p}_{j}, t_{i}^{c}] \Bigg)^2 \\
    &\leq \Bigg(\frac{dm}{dt}(t_i^{c}, \theta_{m_{j}}) 
    - \frac{d\mathbf{y}_{j}}{dt}(t_i^{c}) \Bigg)^2 \\
    &\quad + \Big(f[\mathbf{y}_{j}(t_i^{c}), \mathbf{p}_{j}, t_{i}^{c}] 
    - f[m(t_{i}^{c}; \theta_{m_{j}}), \mathbf{p}_{j}, t_{i}^{c}] \Big)^2 \\
    &\quad + \Bigg(\frac{d\mathbf{y}_{j}}{dt}(t_i^{c}) 
    - f[\mathbf{y}_{j}(t_i^{c}), \mathbf{p}_{j}, t_i^{c}]\Bigg)^2 \\
    &\leq \Bigg(\frac{dm}{dt}(t_i^{c}, \theta_{m_{j}}) 
    - \frac{d\mathbf{y}_{j}}{dt}(t_i^{c}) \Bigg)^2 \\
    &\quad + L_{f}^{2} \Big(\mathbf{y}_{j}(t_i^{c}) 
    - m(t_i^{c}; \theta_{m_{j}})\Big)^2 + 0 \\
    &\leq \frac{\varepsilon}{2(1 + L_{f}^{2})} 
    + \frac{L_{f}^{2} \varepsilon}{2(1 + L_{f}^{2})} = \frac{\varepsilon}{2}.
    \end{align*}
    
    By using the third inequality from Lemma \ref{appendix:lemma_lipschitz} and the above inequality, we can conclude the following: 
      
    \begin{align*}
    L_{\text{physics(disc)}}(\theta_h) &= 
    \frac{1}{T_{col}} \sum_{i=1}^{T_{col}} \frac{1}{N_{p}} 
    \sum_{j=1}^{N_{p}} |\ell(m_{j}(t_i^{c}), h(\mathbf{p}_{j}; \theta_{h}))| \\
    &\leq \frac{1}{T_{col}} \sum_{i=1}^{T_{col}} \frac{1}{N_{p}} 
    \sum_{j=1}^{N_{p}} \Big( |\ell(m_{j}(t_i^{c}), h(\mathbf{p}_{j}; \theta_{h})) \\
    &\quad - \ell(m_{j}(t_i^{c}), \theta_{m_{j}}(\mathbf{p}_{j}))| \\
    &\quad + |\ell(m_{j}(t_i^{c}), \theta_{m_{j}}(\mathbf{p}_{j}))| \Big) \\
    &\leq \frac{1}{T_{col}} \sum_{i=1}^{T_{col}} \frac{1}{N_{p}} 
    \sum_{j=1}^{N_{p}} \left(\frac{\varepsilon}{2L_{\ell}} \cdot L_{\ell} 
    + \frac{\varepsilon}{2}\right) = \varepsilon.
    \end{align*}

\end{proof}

\subsection{Simulation dataset generation}
For a given DE, we generated a synthetic RCS dataset. Specifically, we first designed underlying distributions of parameters that have $H$ different numbers of peaks as follows. For each peak $\mathbf{p}_{h}^{peak}$ with index $h=1,\dots,H$, $S$ different parameters $\{\mathbf{p}_{h(S-1)+i}\}_{i=1}^{S}$ are sampled from uniform distributions $U((\mathbf{p}_{Low})_{h},(\mathbf{p}_{High})_{h})$ with lower and upper bounds $(\mathbf{p}_{Low})_{h}$ and $(\mathbf{p}_{High})_{h}$, respectively:
$$
\mathbf{p}_{h(S-1)+i} \sim U((\mathbf{p}_{Low})_{h},(\mathbf{p}_{High})_{h}).
$$ 

With the $HS$ different values of parameters $\{\mathbf{p}_{j}\}_{j=1}^{HS}$, we generate solutions of Eq.~(\ref{equation:ODE}) through the DE solver with the same initial value $y_{0}$. We refer to Table~\ref{table:data settings for wgan} for the specific parameter values corresponding to each DE. Thus, we have the training dataset $(t_{i},\{\mathbf{y}(t_{i};\mathbf{p}_{j})\}_{i,j=1}^{T,HS}$, which contains the snapshots for each parameter $\mathbf{p}_{j}$. These datasets are used as observation datasets for training the WGAN.

\newpage
\subsection{Supplemental tables}

\begin{table}[h]\centering
\caption{{\bf Settings for generating synthetic RCS data of the exponential growth model.}}\label{table:synthetic exp}
\begin{tabular*}{0.46\textwidth}{ccccc}\addlinespace
\toprule 
\multirow{2.2}{*}{Experiment} & \multirow{2.2}{*}{\makecell{Distribution \\ Type}} & \multirow{2.2}{*}{\makecell{ Initial \\ Condition}} & \multirow{2.2}{*}{\makecell{Parameter \\ Range }} &
\multicolumn{1}{c}{Peaks}  \\
\cmidrule{5-5} & &  & & $r$  \\ 
\midrule\multirow{6.3}{*}{Exponential}&uni-modal&\multirow{6.3}{*}{$Y=1$}&\multirow{6.3}{*}{$r \in [0.5,3.5]$}& $1$ \\ \cmidrule{2-2}\cmidrule{5-5}  & \multirow{2.1}{*}{bi-modal} & & & $1$\\& & & & $3$\\ \cmidrule{2-2}\cmidrule{5-5}& \multirow{3.2}{*}{tri-modal} & &  & $1$\\& & & & $2$\\& & & & $3$\\
\bottomrule
\end{tabular*}\end{table}

\begin{table}[h]\centering
\caption{{\bf Settings for generating synthetic RCS data of the logistic population model.}}\label{table:synthetic log}
\begin{tabular*}{0.48\textwidth}{@{\extracolsep{\fill}}l@{\hskip 0.2cm}ccc@{\hskip 0.2cm}ccc@{\hskip 0.2cm}ccc}
\toprule 
\multirow{2.2}{*}{Experiment} & \multirow{2.2}{*}{\makecell{Distribution \\ Type}} & \multirow{2.2}{*}{\makecell{ Initial \\ Condition}} & \multirow{2.2}{*}{\makecell{Parameter \\ Range }} &
\multicolumn{2}{c}{Peaks}  \\
\cmidrule{5-6} & &  & & $r$ & $K$  \\ 
\midrule\multirow{6.3}{*}{Logistic}&uni-modal&\multirow{6.3}{*}{$Y=10^{-5}$}&\multirow{6.3}{*}{\makecell{$r \in [1,5]$ \\ \\ $K \in [0.2,1.5]$}}& $2.8$ & $1.0$ \\ \cmidrule{2-2}\cmidrule{5-6}  & \multirow{2.1}{*}{bi-modal} & & & $1.6$ & $0.6$\\& & & & $4.0$ & $1.4$\\ \cmidrule{2-2}\cmidrule{5-6}& \multirow{3.2}{*}{tri-modal} & &  & $1.6$ & $0.6$\\& & & & $4.0$ & $0.9$\\& & & & $2.0$ & $1.3$\\ \bottomrule
\end{tabular*}\end{table}

\newpage

\begin{table}[h]\centering 
\caption{{\bf Settings for generating synthetic RCS data of the Lorenz system.}}
\label{table:synthetic lorenz}
\setlength{\tabcolsep}{2pt} 
\renewcommand{\arraystretch}{1.0} 
\begin{tabular*}{0.48\textwidth}{@{\extracolsep{\fill}}ccccccc}
\toprule 
\multirow{2.2}{*}{Experiment} & \multirow{2.2}{*}{\makecell{Distribution \\ Type}} & \multirow{2.2}{*}{\makecell{Initial \\ Condition}} & \multirow{2.2}{*}{\makecell{Parameter \\ Range}} &
\multicolumn{3}{c}{Peaks}  \\
\cmidrule{5-7} & &  & & $\sigma$ & $\rho$ & $\beta$  \\ 
\midrule
\multirow{6.3}{*}{Lorenz} & uni-modal & \multirow{6.3}{*}{\makecell{$X=4.67$ \\\\ $Y=5.49$ \\\\ $Z=9.06$}} & \multirow{6.3}{*}{\makecell{$\sigma \in [9,11]$ \\\\ $\rho \in [0,28]$ \\\\ $\beta \in [2/3,8/3]$}} 
& $9.50$ & $27.0$ & $5/3$ \\ 
\cmidrule{2-2}\cmidrule{5-7}  
& \multirow{2.1}{*}{bi-modal} & & & $10.5$ & $18.0$ & $1$ \\
& & & & $10.0$ & $24.75$ & $7/3$ \\ 
\cmidrule{2-2}\cmidrule{5-7} 
& \multirow{3.2}{*}{tri-modal} & & & $10.5$ & $18.0$ & $1$ \\
& & & & $9.50$ & $27.0$ & $5/3$ \\
& & & & $11$ & $24.75$ & $7/3$ \\
\bottomrule
\end{tabular*}
\end{table}

\newpage

\begin{table}[h]\centering
\caption{{\bf Settings for network structures.} We provide information on the hyperparameters (number of hidden layers and nodes) for the main and hyper networks in HyperPINN, as well as the hyperparameters for the generator and discriminator in WGAN that we used to develop EIDGM. Additionally, the hyperparameters for DeepONet, which was compared to EIDGM, are also provided.}
\label{table:network settings}
\setlength{\tabcolsep}{1.5pt} 
\renewcommand{\arraystretch}{1.0} 
\begin{tabular*}{0.48\textwidth}{@{\extracolsep{\fill}}ccccccccc}
\toprule 
\multirow{2.2}{*}{Experiment} & \multirow{2.2}{*}{Setting} & \multicolumn{2}{c}{DeepONet} & \multicolumn{2}{c}{HyperPINN} & \multicolumn{2}{c}{WGAN} \\
\cmidrule{3-8} 
& & branch & trunk & hyper & trunk & generator & discriminator \\
\midrule
\multirow{4.2}{*}{Exponential} & architecture & \multicolumn{6}{c}{fully-connected} \\ 
\cmidrule{2-8} 
& width & 128 & 128 & 64 & 32 & 64 & 64 \\ 
\cmidrule{2-8} 
& depth & 3 & 3 & 4 & 4 & 4 & 4 \\ 
\cmidrule{2-8} 
& activation & \multicolumn{6}{c}{tanh} \\ 
\midrule 
\multirow{4.2}{*}{Logistic} & architecture & \multicolumn{6}{c}{fully-connected} \\ 
\cmidrule{2-8} 
& width & 128 & 128 & 64 & 32 & 64 & 64 \\ 
\cmidrule{2-8} 
& depth & 3 & 3 & 4 & 4 & 4 & 4 \\ 
\cmidrule{2-8} 
& activation & \multicolumn{6}{c}{tanh} \\ 
\midrule 
\multirow{4.2}{*}{Lorenz} & architecture & \multicolumn{6}{c}{fully-connected} \\ 
\cmidrule{2-8} 
& width & 128 & 128 & 64 & 64 & 128 & 128 \\ 
\cmidrule{2-8} 
& depth & 3 & 3 & 4 & 4 & 4 & 4 \\ 
\cmidrule{2-8} 
& activation & \multicolumn{6}{c}{tanh} \\ 
\bottomrule
\end{tabular*}
\end{table}

\newpage

\begin{table}[h]\centering
\caption{{\bf Settings for training DeepONet and HyperPINN.} We provide information on the settings related to the optimizer and dataset for training both DeepONet and HyperPINN.}
\begin{tabular*}{0.405\textwidth}{cccc}\addlinespace \toprule Experiment & Setting & DeepONet & HyperPINN \\ 
\midrule\multirow{7.3}{*}{Exponential}& optimizer & \multicolumn{2}{c}{Adam}\\ \cmidrule{2-4} & learning rate & $10^{-4}$ & $5 \times 10^{-5}$ \\ \cmidrule{2-4} & batch size & \multicolumn{2}{c}{$10^{4}$} \\ \cmidrule{2-4}& $N_{p}$ & \multicolumn{2}{c}{$10^{2}$} \\ \cmidrule{2-4}& $T_{obs}$, $T_{col}$ & \multicolumn{2}{c}{$10^{2}$, $10^{2}$} \\ \cmidrule{2-4}& $\alpha$, $\beta$ & \multicolumn{2}{c}{$1$, $10^{-2}$} \\ \cmidrule{2-4}& training epochs & \multicolumn{2}{c}{$10^{4}$} \\ \midrule\multirow{7.3}{*}{Logistic}& optimizer & \multicolumn{2}{c}{Adam}\\ \cmidrule{2-4} & learning rate & $10^{-4}$ & $5 \times 10^{-5}$ \\ \cmidrule{2-4} & batch size & \multicolumn{2}{c}{$10^{4}$} \\ \cmidrule{2-4}& $N_{p}$ & \multicolumn{2}{c}{$2\times 10^{2}$} \\ \cmidrule{2-4}& $T_{obs}$, $T_{col}$ & \multicolumn{2}{c}{$10^{2}$, $10^{2}$} \\ \cmidrule{2-4}& $\alpha$, $\beta$ & \multicolumn{2}{c}{$1$, $10^{-2}$} \\ \cmidrule{2-4}& training epochs & \multicolumn{2}{c}{$10^{4}$} \\ \midrule\multirow{7.3}{*}{Lorenz}& optimizer & \multicolumn{2}{c}{Adam}\\ \cmidrule{2-4} & learning rate & $5\times 10^{-4}$ & $5 \times 10^{-5}$ \\ \cmidrule{2-4} & batch size & \multicolumn{2}{c}{$10^{4}$} \\ \cmidrule{2-4}& $N_{p}$ & \multicolumn{2}{c}{$10^{3}$} \\ \cmidrule{2-4}& $T_{obs}$, $T_{col}$ & \multicolumn{2}{c}{$10^{2}$, $10^{2}$} \\ \cmidrule{2-4}& $\alpha$, $\beta$ & \multicolumn{2}{c}{$1$, $0$} \\ \cmidrule{2-4}& training epochs & \multicolumn{2}{c}{$10^{4}$} \\
\bottomrule
\end{tabular*}
\end{table}

\newpage

\begin{table}[h]\centering
\caption{{\bf Settings for training WGAN.} We provide information on the settings related to the optimizer and dataset for training WGAN.}\label{table:data settings for wgan}
\setlength{\tabcolsep}{2pt} 
\renewcommand{\arraystretch}{1.0} 
\begin{tabular*}{0.48\textwidth}{@{\extracolsep{\fill}}cccccccccc}
\toprule 
\multirow{2.1}{*}{Setting} & \multicolumn{3}{c}{Exponential} & \multicolumn{3}{c}{Logistic} & \multicolumn{3}{c}{Lorenz} \\ 
\cmidrule{2-10} 
& uni & bi & tri & uni & bi & tri & uni & bi & tri \\
\midrule 
Optimizer & \multicolumn{9}{c}{Adam($\beta_{1}=0.0$, $\beta_{2}=0.9$)} \\
\midrule 
Learning rate & \multicolumn{9}{c}{$10^{-4}$} \\
\midrule 
Batch size & \multicolumn{9}{c}{Full-batch} \\
\midrule 
$|Y|$ & $60$ & $120$ & $180$ & $60$ & $120$ & $180$ & $108$ & $216$ & $324$ \\
\midrule 
$|\tilde{Y}|$ & $60$ & $120$ & $180$ & $60$ & $120$ & $180$ & $324$ & $648$ & $972$ \\
\midrule 
Noise dimension & \multicolumn{3}{c}{$16$} & \multicolumn{3}{c}{$16$} & \multicolumn{3}{c}{$32$} \\
\midrule 
Training epochs & \multicolumn{3}{c}{$5\times 10^{4}$} & \multicolumn{3}{c}{$5\times 10^{4}$} & \multicolumn{3}{c}{$10^{5}$} \\ 
\bottomrule
\end{tabular*}
\end{table}

\newpage

\end{document}